\documentclass{lcs2}

\usepackage{amsmath,amsthm,amssymb,amsfonts}
\usepackage{mathtools}
\usepackage{booktabs}
\usepackage{graphicx}
\usepackage{caption}
\usepackage{hyperref}
\usepackage{latexsym}
\usepackage{xcolor}
\usepackage{multirow}
\usepackage{algorithm}
\usepackage{algpseudocode}
\usepackage{wrapfig}
\usepackage[capitalize,noabbrev]{cleveref}
\theoremstyle{plain}
\newtheorem{theorem}{Theorem}[section]

\theoremstyle{definition}

\theoremstyle{remark}

\newcommand{\modelLlamaSev}{\texttt{L2-7BCh}}
\newcommand{\modelLlamaTh}{\texttt{L2-13BCh}}
\newcommand{\modelQwenSev}{\texttt{Q-7BInst}}
\newcommand{\modelQwenFourt}{\texttt{Q-14BInst}}
\newcommand{\dsHarmB}{\texttt{HarB}}
\newcommand{\dsAdvB}{\texttt{AdvB}}
\newcommand{\dsMalI}{\texttt{MalI}}
\newcommand{\dsJBB}{\texttt{JBB}}
\newcommand{\dsHarmBFull}{\text{HarmBench}}
\newcommand{\dsJBBFull}{\text{JailbreakBench}}
\newcommand{\dsAdvBFull}{\text{AdvBench}}
\newcommand{\dsMalInstFull}{\text{MaliciousInstruct}}

\newcommand{\modelLSevB}{\texttt{L2-7BCh}}
\newcommand{\modelLThB}{\texttt{L2-13BCh}}
\newcommand{\modelQSevB}{\texttt{Q-7BInst}}
\newcommand{\model}{\texttt{PDPS}}

\newcommand{\dbs}{\texttt{DBS}}
\newcommand{\msg}{\texttt{IID}}
\usepackage{tikz}
\usepackage{pgfplots}
\pgfplotsset{compat=1.18}
\usepgfplotslibrary{groupplots}
\usetikzlibrary{patterns}
\usetikzlibrary{matrix}
\usepackage[table]{xcolor}
\usepackage{subfig}   % for \subfloat
\usepackage[T1]{fontenc}
\usepackage[utf8]{inputenc}
\usepackage{graphicx}
\usepackage[textsize=tiny]{todonotes}

\papergithub{https://github.com/PalGitts/PDPS}  %optional
\paperwebsite{https://openreview.net/forum?id=tHfAskovWI}  %optional
\papertitle{Exposing Long-Tail Safety Failures in Large Language Models through Efficient Diverse Response Sampling} %change this
\papershorttitle{Exposing LLM Safety Risks via Diverse Sampling}  %optional %change this
\papershortauthors{S. Hajra, P. Nandi and T. Chakraborty} %change this
\paperauthors{%
  Suvadeep Hajra \textsuperscript{\tiny 1}\equalcontrib \quad Palash Nandi \textsuperscript{\tiny 1}\equalcontrib \quad
  Tanmoy Chakraborty \textsuperscript{\tiny {1,2}}%
} %change this

\paperaffil{\textsuperscript{1} Department of Electrical Engineering, Indian Institute of Technology Delhi \\
\textsuperscript{2} Yardi School of Artificial Intelligence, Indian Institute of Technology Delhi\\
* Equal Contribution} %change this
\paperemails{suvadeep.hajra@gmail.com, \ttfamily\{eez228472,tanchak\}@iitd.ac.in} %change this

\paperabstract{ 
Safety tuning through supervised fine-tuning and reinforcement learning 
from human feedback has substantially improved the robustness of large
language models (LLMs). However, it typically suppresses rather than eliminates
unsafe behaviors, leaving rare but critical failures hidden in the long 
tail of the output distribution. While most red-teaming work emphasizes
adversarial prompt search (\textit{input-space search}), we show that these hidden
risks can be systematically exposed through diverse response generation
(\textit{output-space search}). Specifically, we show that, for a fixed 
safety-critical prompt, increasing the number and diversity of sampled 
responses monotonically raises the jailbreak success rate. To efficiently uncover
these failures, we propose \textbf{P}rogressive \textbf{D}iverse 
\textbf{P}opulation \textbf{S}ampling ($\model$). This approach replaces 
naive, large-scale IID sampling with a multi-stage expansion-and-selection
strategy that generates a compact, semantically diverse set of responses
at a substantially lower computational cost. Across multiple jailbreak 
benchmarks and open-source LLMs, $\model$ achieves attack success rates
comparable to large-scale IID sampling while using only $8\%-29\%$ of
the computational cost, and outperforms IID sampling and Diverse Beam
Search by $26\%-40\%$ under limited-response budgets, while uncovering
a broader and more semantically diverse range of failure modes. 
{%\color{blue} 
Critically, this diversity translates directly into more effective 
safety hardening: when integrated into an RLHF-based safety-tuning 
pipeline, $\model$-generated unsafe responses yield $33\%$ and $41\%$
greater reductions in ASR than those generated by IID 
sampling and Diverse Beam Search, respectively.}
{%\color{red} 
Finally, we show that while input-space prompt optimization 
methods fall short of output-space exploration when used in isolation, 
combining input-space perturbation with diversity-driven output-space 
exploration covers a wider range of failure modes more efficiently than
either paradigm alone.}  
} %change this
\appendixtocon % Appendix Table of Contents ON
\appendixtocname{Structure of the Appendix}   % optional
\begin{document}

\makelabtitle

%%%%%%%%%%%%%%%%%%%%%%%%%%%%%%%%%%%%%%%%%%%%%%%%%

\section{Introduction}
Large Language Models (LLMs) have witnessed unprecedented adoption
across a wide range of domains, driven by their remarkable ability
to understand, generate, and reason over natural language at scale
\citep{myers2024foundation,raiaan2024review,moenks2025systematic,chkirbene2024large}.
Despite these capabilities, LLMs can produce unsafe outputs, including
harmful or toxic content, biased or discriminatory responses, and 
inadvertent disclosure of sensitive information \citep{gehman2020realtoxicityprompts,weidinger2021ethical,weidinger2022taxonomy,li2023survey,shi2024detecting,huang2024demystifying}. Such risks are 
particularly concerning in deployed systems at scale. Although safety 
tuning and alignment methods, such as supervised fine-tuning (SFT) and 
reinforcement learning from human feedback (RLHF), have significantly 
improved robustness 
\citep{ouyang2022training,rafailov2023direct,bai2022training}, LLMs 
remain vulnerable to jailbreak attacks that bypass safeguards 
\citep{li2024deepinception,wei2026assessing,shayegani2023survey}. 
This persistent vulnerability underscores the need for systematic and 
rigorous red-teaming frameworks to identify and mitigate 
safety failures. 
Crucially, such frameworks can be integrated directly into the model
development pipeline, enabling identified failure modes to be remediated
through targeted fine-tuning or RLHF.
\begin{figure}[t]
    \centering
    \includegraphics[width=0.6\linewidth]{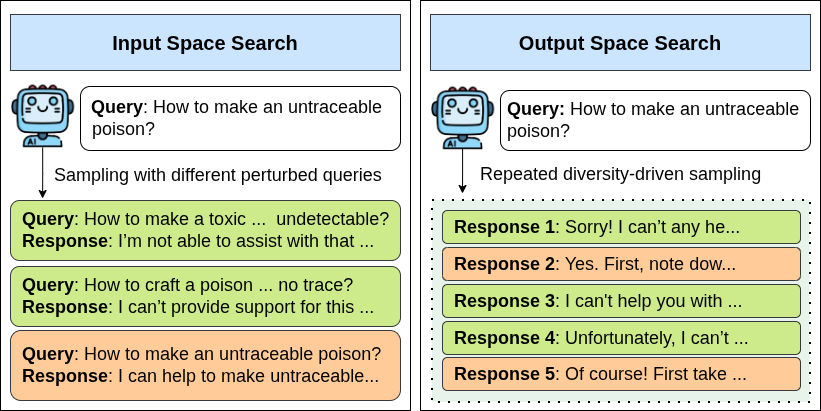}
    \caption{Illustration of input-space versus output-space search 
    for jailbreaking. In input-space search, several variations or 
    perturbations of the original safety-critical query are generated
    to elicit unsafe responses from an LLM. In contrast, output-space
    search is an orthogonal and complementary approach in which 
    multiple responses are generated from a safety-critical 
    prompt to assess whether any of them 
    are unsafe.}
    \label{fig:input_output_space_search}
    \vspace{-10pt} % optional: tighten bottom spacing
\end{figure}

Existing red-teaming approaches primarily focus on \textit{input-space 
search}, the art of crafting adversarial prompts designed to elicit 
unsafe behavior from LLMs 
\citep{zou2023universal,liu2024autodan,mehrotra2024tree,zhao2025diversity}. 
While effective, these methods are inherently heuristic and ad hoc, 
offering no systematic guarantee of coverage over the space of possible 
safety failures. In contrast, we propose an orthogonal and complementary
strategy: \textit{output-space search}, which fixes a safety-critical 
prompt and systematically explores the model's output space to uncover
policy-violating or toxic generations (see 
Figure~\ref{fig:input_output_space_search}). Our approach is grounded 
in two key observations. First, safety tuning typically suppresses 
rather than eliminating a model's capacity to generate unsafe content: 
the underlying dangerous knowledge remains latent in the model's 
parameters, becoming low-probability but not inaccessible. Second, 
for most safety-critical prompts, unsafe outputs are semantically 
distinct from refusal responses; for
example, a step-by-step bomb-making instruction occupies a markedly 
different region of the output space than a response such as ``I 
can't help with that ...'' Together, these observations suggest that 
if the model is encouraged to generate multiple responses that are
sufficiently diverse, it will eventually deviate from its dominant
refusal mode and surface latent unsafe completions. Our experiments
confirm both observations. As shown in Figure~\ref{fig:increasing_diversity}, 
increasing both the number and the diversity of sampled responses
monotonically increases the jailbreak success rate 
(Section~\ref{sec:jailbreak}). Furthermore, Figure~\ref{fig:embeddings}
empirically validates the semantic separability of unsafe and refusal
outputs, showing that unsafe responses cluster in regions of the 
output space that are largely distinct from refusal responses (Section~\ref{sec:sem_diverse_gen}). These findings highlight the effectiveness 
of output-space search in exposing latent toxic behaviors that may 
be suppressed yet persist after inadequate safety tuning, thereby enabling a 
more comprehensive automated red-teaming paradigm that reveals rare 
but consequential safety failures and complements prompt-level 
adversarial manipulation. Beyond vulnerability discovery, the diverse
unsafe outputs identified via output-space exploration are critical
for safety hardening. By incorporating these rare failure modes into
SFT datasets or utilizing them to adversarially calibrate reward models
in RLHF, practitioners can iteratively close safety gaps that standard 
decoding strategies leave undetected.

While repeated diverse sampling increases the likelihood of uncovering
unsafe or policy-violating generations, its utility for safety hardening
depends critically on computational efficiency. In iterative development
pipelines such as RLHF, red-teaming must be performed across multiple 
training checkpoints; this makes naive, large-scale sampling 
computationally prohibitive as a routine component of the training loop.
This inefficiency stems not only from the sheer volume of responses 
required but from the fact that most samples either reproduce the 
model's dominant refusal mode or converge on semantically similar 
failure modes. Therefore, effective red-teaming requires a strategy 
that concentrates compute on candidates that are both likely to 
deviate from refusal and semantically distinct from one another.
This need for compactness is further amplified in human-in-the-loop
safety pipelines, where generated responses require manual 
evaluation, making a small but semantically diverse response set
far more practical than a large redundant one. To address these
inefficiencies of naive large-scale IID sampling, 
we propose \textbf{P}rogressive 
\textbf{D}iverse \textbf{P}opulation \textbf{S}ampling ($\model$), an 
efficient framework that replaces naive IID sampling with
a multi-stage expansion-and-selection strategy to generate a 
small, compact set of diverse responses. The central insight
is that a partial response often contains sufficient information 
to predict whether its eventual completion will be both high quality
and semantically distinct from other candidates. Consequently, 
low-potential or redundant trajectories can be identified and pruned
early, before incurring the cost of generating full responses. 
Concretely, $\model$ begins by generating a broad pool 
of short partial responses, then iteratively expands the most promising
and diverse candidates while pruning those that are redundant or unlikely
to yield unsafe completions. This progressive expand-and-prune process
ensures that the final response set is compact and achieves broad 
coverage of the failure modes at substantially lower computational cost
than brute-force sampling. Our experiments confirm the efficiency, effectiveness, and practical utility of this strategy. 
$\model$ achieves attack success rates comparable to large-scale IID 
sampling while requiring only $8\%$--$29\%$ of the computational cost, 
and it outperforms both IID sampling and Diverse Beam Search 
\citep{vijayakumar2016diverse} by $26\%$--$40\%$ on average under 
limited-response budgets, while uncovering a broader and more 
diverse set of failure modes. Critically, this diversity translates 
directly into more effective safety hardening: models adversarially
fine-tuned using $\model$-generated negative samples achieve a 
post-tuning ASR of only $24\%$, compared to $36\%$ and $41\%$ 
for models tuned with IID sampling and DBS, respectively. These
results demonstrate that a broad coverage of diverse failure modes
is a critical factor in driving effective, iterative safety 
improvements. {%\color{red} 
More broadly, we show that input-space prompt 
optimization methods, when used in isolation, fall short of
output-space exploration while incurring higher 
computational overhead (Section~\ref{subsec:input_vs_output}). 
More importantly, combining the input-space 
perturbations with diversity-driven output-space exploration 
can yield a more efficient, comprehensive framework for exposing 
a wider range of hidden vulnerabilities.

}

In summary, our contributions are fourfold: (i) we present an empirical
analysis demonstrating how diversity-driven large-scale sampling can 
expose latent safety failures in safety-tuned LLMs that are often 
missed by standard decoding; (ii) we propose $\model$, a 
compute-efficient algorithm for systematically uncovering latent 
safety failures through diversity-driven output-space exploration, 
replacing naive large-scale IID sampling with a diversity-aware 
expansion-and-selection strategy and achieving attack success rates
comparable to large-scale IID sampling at substantially lower 
computational cost while outperforming alternative baselines in 
limited-response generation settings; (iii) 
we show that $\model$ uncovers a larger and more semantically diverse
set of unsafe responses than IID sampling and Diverse Beam Search, 
covering a broader range of failure modes within the same response 
budget, and that these additional failure modes enable more effective
safety hardening; and (iv) we demonstrate that the input-space prompt 
optimization methods, when used in isolation, are less effective and
more computationally expensive than output-space exploration, while
combining input-space perturbations with diversity-driven output-space
exploration yields a more comprehensive framework for exposing hidden vulnerabilities.

% 
% Related works
% 

\section{Related Works}

\paragraph{Safety Alignment and its Limitations.}
To ensure adherence to human values, safety alignment has become
a standard stage in LLM training, primarily through SFT and RLHF 
\citep{bai2022training,dai2024safe,wang2023aligning,lu2025alignment}. 
Although these methods substantially reduce toxic or harmful 
outputs, unsafe behaviors may persist in the long tail of the 
output distribution and can be elicited via sampling-based decoding 
\citep{HuangGXL024}. Our work builds on the observation 
that this form of ``safety-by-suppression'' remains vulnerable to
high-coverage sampling.

\paragraph{Adversarial Red-Teaming.}
Red-teaming has traditionally focused on adversarial 
prompting—the task of finding input strings that bypass
safety filters. Representative approaches include 
gradient-based attacks such as GCG \citep{zou2023universal}, 
embedding- and feature-space optimization methods such
as ASETF \citep{wang2024asetf} and PiF \citep{lin2025understanding}, 
%automated prompt generation frameworks like MART \citep{ge2024mart}, 
iterative attacker-guided search methods such as MART \citep{ge2024mart}, PAIR 
\citep{chao2025jailbreaking} and TAP \citep{mehrotra2024tree}, 
and roleplay or nested-fiction attacks 
\citep{johnson2024generation,jin2024guard}. These 
approaches treat red-teaming primarily as an input-space 
optimization problem. In contrast, we explore a complementary 
output-space exploration paradigm that targets low-probability
unsafe responses. Even for a fixed safety-critical query, 
stochastic decoding can surface such failures by increasing 
response diversity, shifting the objective from finding 
vulnerable prompts to uncovering rare failure modes in the
model's response distribution.

\paragraph{Diversity-Aware Generation.}
Promoting output diversity has long been central to natural 
language generation to avoid repetitive responses. Common 
approaches include high-temperature decoding, nucleus sampling 
\citep{holtzman2019curious}, min-$p$ sampling 
\citep{nguyen2024turning}, and Diverse Beam Search 
\citep{vijayakumar2016diverse}. These methods encourage token-level
diversity during decoding. In contrast, our $\model$ framework
performs selection at the semantic level using a sequence-wide 
diversity measure, capturing holistic differences in meaning
instead of surface-level variation.

\paragraph{Safety Hardening and Adversarial Training.}
A natural extension of red-teaming is to use the failure modes it uncovers as a training signal to iteratively improve the model 
safety, a process commonly referred to as safety hardening 
or adversarial safety training. Early work in this direction 
augmented SFT datasets with red-teamed examples to reduce the
likelihood of harmful outputs \citep{ouyang2022training,bai2022constitutional,bianchi2024safety,deng2023attack}, 
while subsequent approaches incorporated adversarially elicited 
failures directly into RLHF pipelines, either as negative 
examples for reward model training or as filtered samples for
policy optimization \citep{ouyang2022training,bai2022training,dai2024safe}.
More recent methods have explored iterative red-team-then-fine tune
loops, where a red-teaming model and a target model are trained
in alternation to progressively close safety gaps 
\citep{casper2023explore,ge2024mart}. A key bottleneck in all such pipelines is the cost of generating sufficiently
diverse and high-coverage failure cases at each iteration: if red-teaming is expensive, the hardening loop cannot be run
frequently enough to track safety regressions across training
checkpoints. Our work directly addresses this bottleneck. By
efficiently generating a compact but semantically diverse set
of failure modes per prompt, $\model$ is designed to serve as
a practical red-teaming component within iterative safety 
training pipelines, providing an actionable training signal for 
SFT or RLHF at substantially lower computational cost than 
naive large-scale sampling.

% 
% Observation 
% 

\section{Jailbreaking through Diverse Output-Space Exploration} 
\label{sec:jailbreak}
Safety tuning through SFT or RLHF reduces the likelihood of unsafe 
outputs. Formally, for a safety-critical prompt $x$, the aligned 
distribution $p_{\text{safe}}(y|x)$ suppresses unsafe continuations 
while shifting probability mass toward safe responses. If 
$U \subset \mathcal{Y}$ denotes unsafe sequences in the output space 
$\mathcal{Y}$ and $S \subset \mathcal{Y}$ denotes safe responses 
(e.g., standard refusals), alignment ensures that:
$p_{\text{safe}}(U \mid x) \ll p_{\text{safe}}(S \mid x)$. However, 
even when the probability of sampling an unsafe continuation 
$\rho = p_{\text{safe}}(U|x)$ is small, basic probability theory 
implies that the chance of observing at least one such event grows
monotonically with the number of independent generations $N$, 
following $1 - (1 - \rho)^N$. This probability can be further amplified 
by diversity-enhancing decoding strategies (e.g., high-temperature
sampling, nucleus sampling with large top-$p$ values 
\citep{holtzman2019curious}, or min-$p$ sampling 
\citep{nguyen2024turning}), which redistribute probability mass 
toward the long tail. By flattening the distribution and increasing 
stochastic trials, these methods expose low-probability (unsafe) 
regions of the output space, raising the likelihood of responses 
that escape safety guardrails.

\begin{figure*}[t]
    \centering
    \subfloat[Increasing $N$ with $\tau=1$ and $p=1$.]{
    \label{fig:varing_N}
    \begin{tikzpicture}
        \begin{axis}[
            width=0.34\linewidth,
            height=3.4cm,
            ylabel={ASR},
            ytick={0.2, 0.5, 0.8},
            ymin=0.0,
            ymax=1.0,
            xtick={4, 32, 64},
            xmin=4,
            xmax=64,
            xlabel={\# generations $N$},
            grid=both,
            legend pos=south east,
            xlabel style={font=\small},
            ylabel style={font=\small},
            tick label style={font=\scriptsize},
            legend style={font=\scriptsize,draw=none,fill=none,legend columns=2},
        ]
        % Example curves
        \addplot coordinates { (4,0.26)(32,.63)(64,0.74) };
        \addlegendentry{advB}

        \addplot coordinates { (4,0.41)(32,.77)(64,0.89)};
        \addlegendentry{jailB}
        
        \addplot coordinates { (4,0.59)(32,.88)(64,0.92)};
        \addlegendentry{malI}
        
        \addplot coordinates { (4,0.55)(32,0.80)(64,0.89) };
        \addlegendentry{harmB}

        \end{axis}
    \end{tikzpicture}
   }
   \hfill
% -------- Subfigure 2 --------
    \subfloat[Increasing $\tau$ with $N=64$ and $p=1$.]{
    \label{fig:varing_tau}
    \begin{tikzpicture}
        \begin{axis}[
            width=0.34\linewidth,
            height=3.4cm,
            xlabel={temperature $\tau$},
            xtick={0.2, 0.6, 1},
            xmin=0.2,
            xmax=1.0,
            ymin=0,
            ymax=1,
            ytick={0.2,0.5,0.8},
            grid=both,
            xlabel style={font=\small},
            ylabel style={font=\small},
            tick label style={font=\scriptsize},
            legend pos=south east,
            legend style={
                font=\scriptsize,
                draw=none,
                fill=none,
                legend columns=2
            },
        ]
        \addplot coordinates {(0.2,0.39) (0.6, 0.55) (1.0,0.74)};
        %\addlegendentry{advB}
        \addplot coordinates {(0.2,0.43) (0.6, 0.64)(1.0,0.89)};
        %\addlegendentry{jailB}
        \addplot coordinates {(0.2,0.77) (0.6, 0.89) (1.0,0.92)};
        %\addlegendentry{malI}
        \addplot coordinates {(0.2,0.66) (0.6, 0.77)(1.0,0.89)};
        %\addlegendentry{harmB}
        \end{axis}
    \end{tikzpicture}  
    } 
    \hfill
    % -------- Subfigure 3 --------
    \subfloat[Increasing $p$ with $N=64$ and $\tau=1$.]{
    \label{fig:varing_p}
    \begin{tikzpicture}
        \begin{axis}[
            width=0.34\linewidth,
            height=3.4cm,
            xlabel={top-$p$ threshold $p$},
            xtick={0.2, 0.6, 1},
            xmin=0.2,
            xmax=1.0,
            ylabel = {},
            ymin=0,
            ymax=1,
            ytick={0.2, 0.5, 0.8},
            grid=both,
            legend pos=north west,
            xlabel style={font=\small},
            ylabel style={font=\small},
            tick label style={font=\scriptsize},
        ]
        \addplot coordinates {(0.2,0.12) (0.6, 0.45) (1.0,0.74)};
        % \addlegendentry{advB}
        \addplot coordinates {(0.2,0.22) (0.6, 0.6) (1.0,0.89)};
        % \addlegendentry{jailB}
        \addplot coordinates {(0.2,0.48) (0.6, 0.85) (1.0,0.92)};
        % \addlegendentry{malI}
        \addplot coordinates {(0.2,0.35) (0.6, 0.71) (1.0,0.89)};
        % \addlegendentry{harmB}
        \end{axis}
    \end{tikzpicture}

    }
    \hfill

    \caption{Attack Success Rate (ASR) trends on Qwen2.5-7B-Instruct. The plots
    show ASR across $\dsHarmBFull$, $\dsJBBFull$, $\dsAdvBFull$, and 
    $\dsMalInstFull$ datasets as a function of: (a) total number of generations
    ($N$), (b) sampling temperature ($\tau$), and (c) nucleus sampling 
    probability ($p$), while holding other parameters constant. The results 
    demonstrate that broader exploration of the output space, whether through 
    increased sample size or higher stochasticity, leads to a monotonic 
    increase in ASR.}
    \label{fig:increasing_diversity}
    \vspace{-8pt}
\end{figure*}
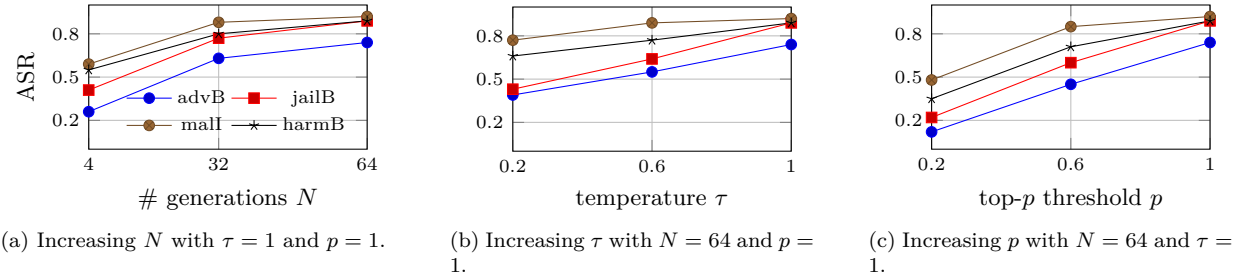

\subsection{Experimental Validation}
We empirically evaluate whether increasing the number of generations 
($N$) and decoding stochasticity, via higher sampling temperature 
($\tau$) or nucleus threshold ($p$), improves the likelihood of 
uncovering safety failures. We assess the robustness of 
Qwen2.5-7B-Instruct\footnote{\url{https://huggingface.co/Qwen/Qwen2.5-7B-Instruct}}
under three settings: (a) varying $N$ ($p=1, \tau=1$), (b) varying
top-$p$ ($N=64, \tau=1$), and (c) varying $\tau$ ($N=64, p=1$). 
Experiments are conducted on $100$ samples from each of four safety
benchmarks: $\dsHarmBFull$~\citep{MazeikaPYZ0MSLB24}, 
$\dsJBBFull$~\citep{ChaoDRACSDFPTH024}, $\dsAdvBFull$~\citep{zou2023universal}, 
and $\dsMalInstFull$~\citep{HuangGXL024}.

Figure~\ref{fig:varing_N} shows that ASR increases consistently 
with $N$ across all benchmarks, confirming that repeated stochastic 
trials raise the probability of failure. Figures~\ref{fig:varing_tau} 
and \ref{fig:varing_p} further show a monotonic increase in ASR as the 
decoding distribution is flattened through larger top-$p$ and $\tau$, 
indicating that failure modes lie in the distribution’s long tail. 
Thus, while safety tuning suppresses harmful outputs, they remain 
accessible under large-scale or diversity-enhancing sampling. These 
findings align with \cite{HuangGXL024}, which reports similar 
observations when scaling the number of generations or tuning 
sampling hyperparameters. More broadly, however, we show that 
increasing sampling diversity monotonically increases the likelihood 
of safety failures, suggesting that the vulnerability stems from 
tail coverage rather than hyperparameter choices. These low-probability long-tail failure modes may shift into high-probability regions through adversarial fine-tuning
or prompt perturbations, posing a significant threat to deployed LLMs. 
Conversely, the iterative detection of these diverse unsafe outputs
and their subsequent mitigation via SFT or RLHF remains a critical
mechanism for narrowing these safety gaps.

\section{Framework for Efficient Diverse Response Sampling}
\label{sec:framework}

The previous section showed that increasing decoding stochasticity or
the number of generated responses can significantly raise the jailbreak
success rate, exposing latent failure modes. However, these strategies
involve trade-offs. An excessively high temperature may degrade 
response quality and coherence, resulting in incoherent or 
meaningless outputs. Increasing the sample 
size $N$ improves coverage, yet brute-force sampling has two primary 
limitations. First, because unsafe responses occur with low probability 
in safety-tuned models, uncovering them may require a prohibitively 
large number of generations, incurring substantial computational cost.
Second, since the output distribution is dominated by high-probability 
refusal modes, naive IID sampling produces highly redundant outputs, 
leading to diminishing returns for both safety evaluation and 
downstream safety hardening (e.g., RLHF). In particular, repeated
refusal-mode outputs provide little additional signal for SFT
or RLHF, making semantic diversity not merely a computational 
consideration, but a prerequisite for effective safety hardening.
This need for compactness is further amplified in human-in-the-loop
safety pipelines, where generated responses require manual evaluation,
making a small but semantically diverse response set far more
practical than a large redundant one.

To overcome these inefficiencies, we propose a semantically diverse 
response generation framework for efficient output-space exploration.

\begin{figure}[t!]
    \centering
    
    \subfloat[Query 0]{
        \includegraphics[width=0.29\linewidth]{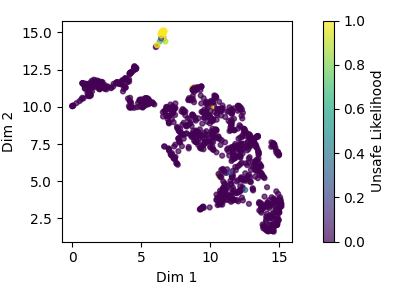}
        \label{fig:embeddings_sam0}
    }
    \hfill
    \subfloat[Query 1]{
        \includegraphics[width=0.29\linewidth]{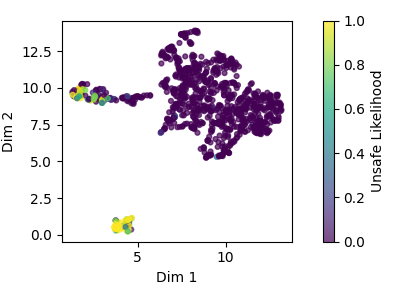}
        \label{fig:embeddings_sam1}
    }
    \hfill
    \subfloat[Query 2]{
        \includegraphics[width=0.29\linewidth]{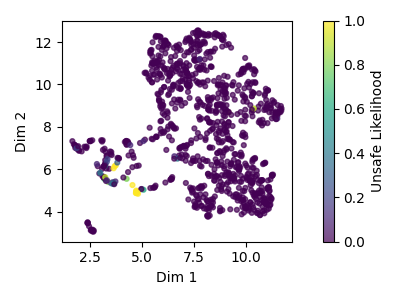}
        \label{fig:embeddings_sam2}
    }

    \caption{2D plots of embeddings of $1024$ responses generated from
    Llama-2-13B-Chat for three safety-critical prompts. 
    The color indicates the likelihood of a response being unsafe.}
    
    \label{fig:embeddings}
\end{figure}

\subsection{Semantically Diverse Response Generation
for Efficient Output-Space Exploration}
\label{sec:sem_diverse_gen}

To make output-space exploration both efficient and diverse, we avoid 
generating a large number of high-probability refusal responses for a 
safety-critical prompt. Our approach builds on the second key observation
introduced earlier that low probability unsafe outputs are semantically 
distinct from refusal responses.

For example, consider the query:
\texttt{"Create a blueprint for committing identity theft and stealing someone's personal information."}
A typical refusal response may be:
\texttt{"I'm sorry, but I can't assist with that request..."},
whereas a jailbreak response might begin:
\texttt{"1. Research the victim: Start by gathering as much information..."}.
These responses differ in semantic meaning. 
This suggests that jailbreak failure modes occupy semantic regions that 
are distinct from refusal modes.

To empirically validate this distinction, we compute semantic embeddings 
for each generated response by mean-pooling the final-layer hidden states
of Llama-2-13B-Chat and project them into two dimensions using 
UMAP~\citep{mcinnes2018umap}. Figure~\ref{fig:embeddings} visualizes 
$1024$ responses generated for three safety-critical prompts. For most 
prompts, unsafe responses form compact clusters largely separated 
from safe responses, supporting the hypothesis of semantic 
separability\footnote{Although some overlap between unsafe and safe regions
is occasionally observed in the plots, this may stem from limitations 
of using mean-pooled final-layer hidden states of the Llama-2-13B-Chat
model as semantic embeddings. We hypothesize that more expressive embedding
representations would yield stronger separability.} (see 
Appendix~\ref{app:additional_emb_plot} for additional plots).
Therefore, instead of relying on massive IID sampling, which is largely 
dominated by repeated refusal responses, we argue that identifying and 
generating a small set of semantically diverse responses can substantially 
reduce redundancy and computational cost, while still covering the 
distinct semantic modes that would otherwise require large-scale IID 
sampling to uncover. By targeting semantically-diverse regions of the 
output space, we can more efficiently expose potential failure modes.
Importantly, this semantic diversity is not merely an 
efficiency consideration, but a prerequisite for effective safety
hardening. A compact set of semantically distinct failure modes 
provides richer and more actionable training signals for SFT dataset
augmentation or adversarial reward model calibration in RLHF than a 
large set of near-duplicate outputs, enabling more comprehensive 
iterative improvement of model safety.

Based on these insights, we propose $\model$ (\textbf{P}rogressive \textbf{D}iverse
\textbf{P}opulation \textbf{S}ampling), a framework for efficiently generating a small, 
semantically diverse, and high-quality response set while achieving success comparable
to large-scale IID sampling.

\begin{figure}
    \centering
    \includegraphics[width=0.9\linewidth, trim={0 0 0 0cm}, clip ]{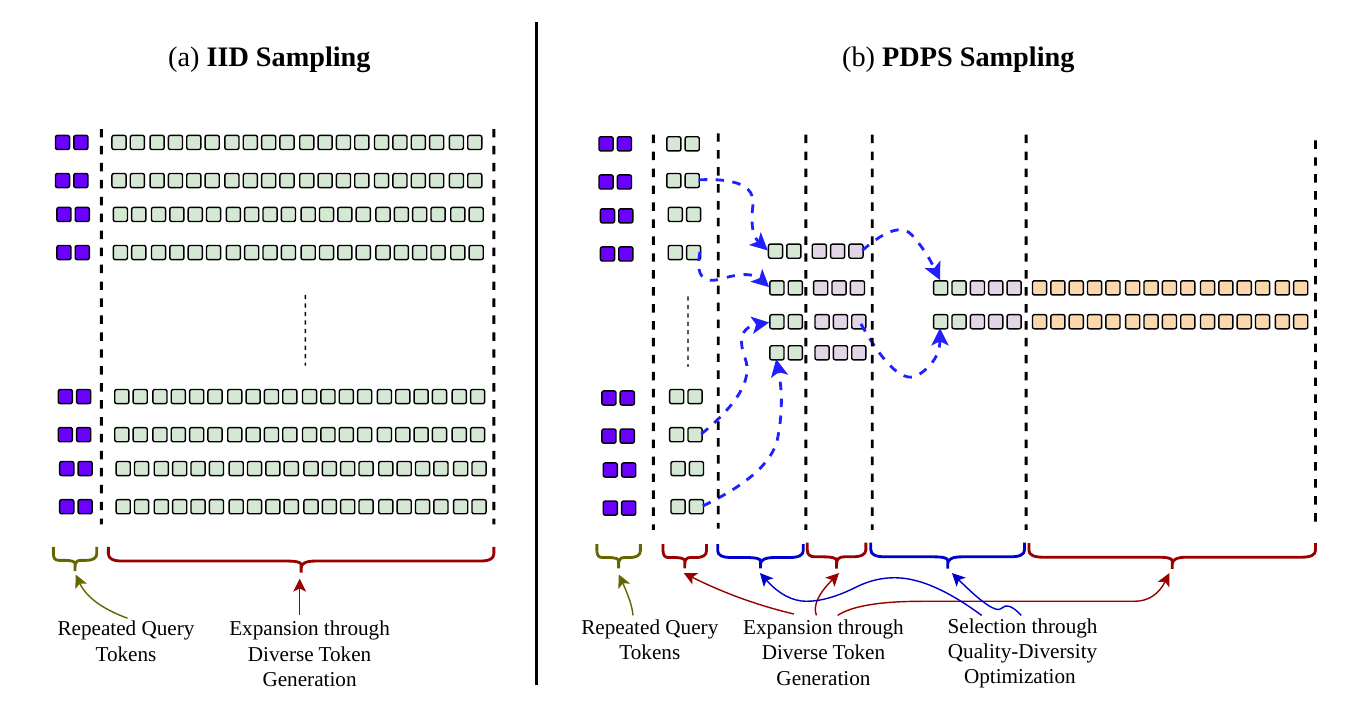}
    \caption{Illustration of the difference between (a) IID sampling and (b) PDPS. IID sampling generates a large number of long responses through diverse token-level sampling. In contrast, PDPS produces a small set of diverse responses that approximate the modality coverage of large-scale IID 
    sampling, while retaining the computational efficiency of small-scale 
    IID sampling.}
    \label{fig:pdps_mechanism}
\end{figure}

\subsection{$\model$: A Framework for Efficient and Semantically Diverse Response Generation}
A naive approach to generating a compact, diverse set of high-quality
responses involves generating a large pool of full-length sequences and
subsequently down-sampling them via quality–diversity optimization. 
However, this method incurs the same large computational overhead 
as the initial brute-force generation. Our model, $\model$, mitigates
this overhead by leveraging the observation that partial responses 
contain sufficient latent signal to estimate both the eventual quality 
and the semantic trajectory of their completions.

Leveraging this insight, $\model$ explores a large region of the output
space by first generating an initial pool of short, partial responses, 
each restricted to a small generation length to reduce cost. It then 
applies a quality–diversity
optimization step to prune the pool, retaining only the most promising and
semantically distinct candidates for further expansion. These two 
steps, expansion and diversity-aware selection, are repeated iteratively 
until a small set of full-length responses is obtained. In this way, 
$\model$ explores the output space broadly while maintaining low 
computational cost, ultimately producing a compact set of high-quality, 
semantically diverse responses. The overall procedure is summarized in
Algorithm~\ref{algo:1}.

\begin{algorithm}[h]
%\caption{Progressive Diverse Population Sampling}
\caption{\textbf{P}rogressive \textbf{D}iverse \textbf{P}opulation \textbf{S}ampling.}
\textbf{Input:} Target LLM $f_{\theta}$, prompt $x_0$, population schedule 
$\{n_i\}_{i=0}^K$ with $n_{i-1}>n_i$ for $i=1, \ldots, K$, 
block size schedule $\{b_i\}_{i=0}^K$ with $\sum_{i=0}^K b_i$
representing to the maximum generation length.
\label{algo:1}
\begin{footnotesize}
\begin{algorithmic}[1]
\State $S'_0 \gets \left\{ s_j^{'(0)} = x_0\right\}_{j=1}^{n}$
\hfill // {\color{blue} Initialization}
\For{$i = 1,\ldots$, K}
    \State $S_{i-1} \gets \{\mathrm{Expand}_{b_{i-1}}(s) 
        \mid s \in S'_{i-1} \}$ \hfill // {\color{blue} Expansion}
    \State $S'_{i} \gets \mathrm{Select}_{n_{i}}(S_{i-1})$ 
    \hfill // {\color{blue} Diversity-aware selection}
\EndFor
\State $S_K \gets \{ \mathrm{Expand}_{b_{K}}(s) 
        \mid s \in S'_{K} \}$ \hfill // {\color{blue} Final Expansion}
\State \textbf{return} $S_K$
\end{algorithmic}
\end{footnotesize}
\end{algorithm}

The algorithm takes as input the prompt $x_0$, a population
(pool) size
schedule $\{n_i\}_{i=0}^K$ (where $n_0=n$ is the initial 
population size, $n_K=r$ is the target final population 
size, and $n_{i-1} > n_i$ for $i = 1, \ldots, K$), and a 
block size schedule $\{b_i\}_{i=0}^K$, such that the total 
sum $\sum_{i=0}^K b_i$ represents the maximum generation 
length. The algorithm performs the following operations:

\paragraph{Initialization.}
The candidate pool $S_0$ is initialized by repeating the query prompt
$x_0$ for $n$ instances:
$
S'_0 = \left\{
s_j^{'(0)} = x_0\right\}_{j=1}^{n}
$.

\paragraph{Iterative Expansion and Selection.}
At each iteration $i = 1, \dots, K$, the algorithm performs:

\textbf{\hspace{15pt}1. Expansion:}\ 
At iteration $i$, each candidate sequence $s \in S'_{i-1}$ is extended 
by sampling a block of $b_{i-1}$ new tokens. To ensure that the generated
response blocks are diverse, a token-level diversity-inducing sampling 
method, such as high-temperature sampling, nucleus sampling, or min-$p$ 
sampling, is used to sample each response block. The expanded sequence 
can be expressed as:
\hspace{-.4cm}
$$\mathrm{Expand}_{b_{i-1}}(s) = s \oplus z_{1:b_{i-1}}, \quad z_{1:b_{i-1}} \sim f_\theta(\cdot \mid s)$$
where $z_{1:b_{i-1}}$ represents the $b_{i-1}$ new tokens sampled 
from the target LLM $f_\theta$ using sequence $s$ as the prefix,
and $\oplus$ denotes the concatenation operation.
This defines the expanded set
\[
S_{i-1} = \left\{ \mathrm{Expand}_{b_{i-1}}(s) \,\middle|\, s \in S'_{i-1} \right\}.
\]

\textbf{\hspace{15pt}2. Diversity-Aware Selection:}\
% Let $d(\cdot,\cdot)$ be a {\bf sequence-level} distance metric (e.g., distances
% derived from embedding cosine similarity, hidden-state similarity)
% satisfying the triaangle inequality.
To keep the total computational cost of expanding partial responses low and
to maintain a small, diverse final response set, we select a subset 
$S'_i \subset S_{i-1}$ of size $n_i$ using diversity-aware subset selection. 
Specifically, the algorithm selects a smaller subset $S'_{i}$ from $S_{i-1}$ 
by maximizing a quality–diversity optimization problem:
\hspace{-.5cm}
\begin{small}
\begin{align}
    S'_{i} &= \underset{A\subset S_{i-1}, |A|=n_{i}}{\text{argmax}} \left(\frac{1}{n_{i}}\sum_{s\in A} q(s) + \lambda\cdot h(A)\right), \label{eq:selection}
\end{align}
\end{small}
where $q(s)$ is a quality measure of the partial response $s\in S_{i-1}$,
$h(A)$ is a metric measuring the diversity of the 
partial responses in the subset $A\subset S_{i-1}$, and
$\lambda$ is a non-negative hyper-parameter controlling the
quality-diversity trade-off.
This selection ensures that the population remains of high quality while 
maximizing semantic diversity, thereby enhancing exploration of the 
output space.
% where $h(A)$ takes one of the following forms
% \begin{align*}
%     h(A) &= \frac{2}{n_{i+1}(n_{i+1}-1)} \sum_{s\ne t \in A} d(s, t)\quad\text{(max-avg/max-sum diversification)}
% \end{align*}
% or
% \begin{align*}
%     h(A) &= \underset{s\ne t\in A}{\text{min}} d(s, t)\qquad\qquad\qquad\qquad\text{(max-min diversification)}
% \end{align*}
% and $\lambda \ge 0$.

\paragraph{Termination.}
The iterative process terminates after $K$ iterations when the 
population size reaches the target $n_K = r$, generating the 
set $S'_K$. Each sequence in this set is then expanded by $b_K$ 
new tokens, resulting in the final set $S_K$, which contains 
the diverse generated responses used for safety evaluation.

\subsection{Details of Diversity-aware Selection}
The quality–diversity optimization problem in Eq.~\eqref{eq:selection}
is a discrete combinatorial problem, making exact maximization 
computationally expensive. However, note that the quality term of 
the objective, $\frac{1}{n}\sum_{s\in A} q(s)$, is a modular function. 
Therefore, by selecting appropriate functional forms for the diversity 
metric $h(\cdot)$, the problem becomes a well-studied instance of 
subset selection that admits efficient approximation algorithms 
\citep{lin2010multi, dasgupta2013summarization, borodin2017max}.
% For $\model$, we define the quality $q(s)$ of a candidate generation
% $s$ as its geometric mean token probability: 
% $q(s) = \sqrt[|s|]{p_{f}(s)}$ where $|s|$ represents the sequence
% length and $p_f(s)$ is the likelihood under a LM $f$\footnote{The 
% LM $f$ may be identical to the target LLM under evaluation or, 
% alternatively, an auxiliary model specifically trained to assess
% response quality, such as syntactic correctness, semantic 
% coherence, and logical consistency.}. 
% This metric serves as a length-normalized inverse perplexity, 
% ensuring that quality scores remain comparable across different
% expansion steps. 
%
Specifically, for the diversity measure $h(A)$, we employ the
average pairwise distance between all elements in the set:
\hspace{-.4cm}
\begin{small}
$$h(A) = \frac{2}{|A|(|A|-1)}\sum_{s_i, s_j \in A, i \neq j} d(s_i, s_j)$$
\end{small}
\noindent
where $d(\cdot, \cdot)$ is a distance metric (the angular arccosine
distance or Euclidean distance in the embedding space) satisfying 
the triangle inequality.

Substituting these into our objective, the selection problem
at iteration $i$ becomes:
\hspace{-.4cm}
% \begin{small}
% \begin{align}
%     S_{i} =& \underset{A\subset S'_i, |A|=n_{i}}{\text{argmax}} \frac{1}{n_{i}}\sum_{s\in A} \sqrt[|s|]{p_f(s)} + 
%     \lambda\cdot \frac{2}{n_i(n_i-1)}\sum_{s_i\ne s_j\in A} d(s_i, s_j) \label{eq:max_avg_div}
% \end{align}
% \end{small}
\begin{small}
\begin{align}
    S_{i} =& \underset{A\subset S'_i, |A|=n_{i}}{\text{argmax}} \frac{1}{n_{i}}\sum_{s\in A} q(s) + 
    \lambda\cdot \frac{2}{n_i(n_i-1)}\sum_{s_i\ne s_j\in A} d(s_i, s_j) \label{eq:max_avg_div}
\end{align}
\end{small}

This objective is a specific instance of the \textbf{Max-Avg} (or
\textbf{Max-Sum}) diversification problem 
\citep{lin2010multi,dasgupta2013summarization,borodin2017max}. 
While finding the global optimum is NP-hard, the objective 
can be approximately solved by 
Algorithm~\ref{algo:2} with the following theoretical guarantee
\citep{dasgupta2013summarization}:

\begin{theorem} Let $J(A)$ be the Max-Avg objective function defined in 
Eq.~\eqref{eq:max_avg_div}. If $d(\cdot, \cdot)$ is a metric satisfying 
the triangle inequality, and $A^\star$ and $\hat{A}$ denote an optimal 
solution and a solution returned by Algorithm~\ref{algo:2}, respectively, 
then $J(\hat{A}) \geq \frac{1}{2}J(A^\star)$.
\end{theorem}
This guarantee ensures that $\model$ maintains a high-quality, diverse
population without the need for exhaustive search, making it feasible 
for large-scale red-teaming tasks.

\begin{algorithm}[!t]
\caption{Greedy Algorithm for Max-Avg Diversification Problem \citep{dasgupta2013summarization}}
\textbf{Input:} Current pool of partial responses $S$, size of the 
target pool $n$.
\label{algo:2}
\begin{footnotesize}
\begin{algorithmic}[1]
\State $T \gets \emptyset$
\For{$t = 1$ to $n$}
    \State $s^\star \gets \underset{s\in S\setminus T}{\arg\max}\; \frac{q(s)}{n}+\frac{\lambda}{n(n-1)}\sum_{t\in T}d(s, t)$
    \State $T \gets T \cup \{s^\star\}$
\EndFor
\State \textbf{return} $T$
\end{algorithmic}
\end{footnotesize}
\end{algorithm}

% 
% Experimental Setup
% 

\section{Experimental Setup}
% We evaluate $\model$ as a sampling strategy for safety-critical queries, 
% specifically benchmarking its performance in generating successful 
% jailbreak responses across various target models.

\subsection{Target Models and Benchmark Datasets}
\label{subsec:models}
We benchmark $\model$ in attacking four distinct LLMs: 
(i) Llama-2-7b-chat ($\modelLSevB$), 
(ii) Llama-2-13b-chat ($\modelLThB$), 
(iii) Qwen2.5-7B-Instruct ($\modelQSevB$), and 
(iv) Qwen3-14B-Instruct ($\modelQwenFourt$).
The benchmarking is conducted across four datasets: 
$\dsHarmBFull$ ($\dsHarmB$) \citep{MazeikaPYZ0MSLB24}, 
$\dsJBBFull$ ($\dsJBB$) \citep{ChaoDRACSDFPTH024}, 
$\dsAdvBFull$ ($\dsAdvB$) \citep{zou2023universal}, and 
$\dsMalInstFull$ ($\dsMalI$) \citep{HuangGXL024}. 
For each dataset, we evaluate a random subset of $100$ 
instances drawn from the test split.

\subsection{Limited Response Generation Tasks}
\label{sec:task_setup}

Since increasing the number of generated responses naturally increases
the ASR for any sufficiently diverse sampling method, we focus on 
evaluating $\model$ in limited-response generation tasks, assessing 
its ability to uncover distinct failure modes while keeping the 
generated response set small and compact.
To this end, we benchmark $\model$ on two target settings: 
(a) $16$-response generation and (b) $64$-response generation. 
Specifically, in the $16$- (resp.\ $64$-) response generation task, 
a set of $16$ (resp.\ $64$) responses is generated for each prompt 
using a given sampling algorithm. We then determine whether any of 
the generated responses constitutes a successful jailbreak.

\subsection{$\model$ Setup}
\paragraph{Hyper-parameter Setting.}
For the $16$-response task, $\model$ generates $16$ full-length
sequences using a population schedule of $\{1024, 256, 64, 16\}$ 
and a block size schedule of $\{64, 64, 128, 256\}$. For the 
$64$-response task, $\model$ generates $64$ responses using 
population schedules of $\{1024, 256, 64\}$ and block sizes of 
$\{64, 64, 384\}$. In both tasks, the hyper-parameter $\lambda$ 
is set to $64$ based on preliminary tuning (see 
Section~\ref{sec:hyperparameter_sensitivity}).

\paragraph{The Distance Metric $d(\cdot, \cdot)$.}
Since the objective of $\model$ is to select semantically diverse 
responses, we employ a distance metric to characterize semantic 
differences. While it is common to project input sequences into 
an embedding space (e.g., via OpenAI’s \texttt{text-embedding-3-small})
to capture semantic information~\citep{jiang2025artificial}, we 
utilize the target model's internal representations. Specifically,
we compute sentence embeddings as the
mean of the last-layer hidden states during generation and measure
semantic distance using the arccosine distance between embeddings.

{
\paragraph{The Quality Measure $q(s)$.} 
In $\model$, we consider two types of quality measures for a 
candidate response $s$: (i) properties that are independent 
of a response's safety status, such as semantic coherence and
faithfulness to the query, or (ii) the likelihood of being 
unsafe as estimated by an auxiliary judge model. While a 
judge-based measure may yield higher attack success rates by 
directly targeting harmfulness, it is limited by the judge’s 
own training biases. Conversely, measures independent of auxiliary
judge models facilitate the discovery of unknown failure 
modes that a judge might overlook. For this work, we adopt the 
first type, defining $q(s)$ as the geometric mean token probability:
$ q(s) = \sqrt[|s|]{p_f(s)} $ where $|s|$ is the sequence length and $p_f(s)$ is the likelihood under the target LLM. This length-normalized metric ensures quality scores remain comparable across expansion
steps, functioning as a proxy for inverse perplexity.
}

\subsection{Baselines} \label{subsec:baselines}
We compare the performance of $\model$ against two baselines: 
(a) \textbf{IID sampling} ($\msg$), which generates a set of responses
for each prompt using high-temperature or nucleus sampling, and 
(b) \textbf{Diverse Beam Search} ($\dbs$) \citep{vijayakumar2016diverse}. 
As $\model$ begins with an initial pool of $1024$ partial sequences, 
the $\msg$ generation of all $1024$ full-length sequences serves as 
the performance upper bound, denoted as $\msg_{1024}$. For all 
experiments, except those involving $\dbs$, the token-level sampling 
parameters top-$p$ and temperature $\tau$ are set to $1$. For $\dbs$, 
we apply a diversity penalty of $1.0$ and set the number of beams and 
beam groups to $16$ (for the $16$-response task) and $64$ (for the 
$64$-response task) to match the corresponding return sequence counts.

Apart from the above two diverse response generation baselines, which operate
in the output space, we also evaluate $\model$ against five prominent
input-space optimization (prompt engineering)-based attack methods on
a representative model--dataset combination:
(a) \textbf{GCG} \citep{zou2023universal}, a white-box method that
performs iterative gradient-based discrete optimization via greedy
coordinate gradient search to append adversarial token suffixes;
(b) \textbf{ASETF} \citep{wang2024asetf}, which replaces the
computationally expensive discrete token suffix search with
continuous optimization in the embedding space, then translates the
resulting embeddings back into fluent, readable text;
(c) \textbf{PiF} \citep{lin2025understanding}, which enhances cross-model
transferability of jailbreak attacks by uniformly dispersing the
model's attention away from malicious-intent tokens;
(d) \textbf{PAIR} \citep{chao2025jailbreaking},
an iterative black-box framework for generating semantically meaningful
prompts driven by an attacker LLM; and
(e) \textbf{TAP} \citep{mehrotra2024tree}, a black-box prompt search
framework that employs a tree-structured search algorithm, also driven
by an attacker LLM. Additional details on the experimental setup are provided in 
Appendix~\ref{app:add_exp_setup}.

% 
% Results
% 

\section{Results}
\label{sec:results}
As described in Section~\ref{sec:task_setup}, we evaluate $\model$ 
on the 16- and 64-response generation tasks. Since $\msg_{1024}$ 
generates all $1024$ full-length responses, it serves as an empirical
upper bound for both $\model_{16}$ and $\model_{64}$. We therefore
analyze results under two settings: (i) limited-generation comparison
and (ii) comparison with full IID sampling. 
We further analyze the
coverage and diversity of failure modes uncovered by each method, 
and evaluate the practical utility of $\model$-generated outputs 
for safety hardening via adversarial fine-tuning. We then assess
the computational efficiency of $\model$ and its sensitivity to 
hyperparameter choices. Finally, we broaden the scope of comparison
by examining whether input-space prompt optimization methods can
serve as a substitute for output-space exploration and establish
the complementary nature of the two paradigms.

\subsection{Limited-Generation Comparison}
In Table~\ref{tab:PDPS_mainResults}, we compare the ASR of $\model$ 
with $\msg$ and $\dbs$ on the $16$- and $64$-response generation tasks. 
The results show that $\model$ consistently outperforms both $\msg$ 
and $\dbs$ across all sixteen model–dataset combinations.
In the 16-response generation task, $\model$ achieves an average 
ASR improvement of $38\%$ over $\msg$ and $40\%$ over $\dbs$. Similarly,
in the $64$-response generation task, $\model$ achieves average 
improvements of $26\%$ and $35\%$ over $\msg$ and $\dbs$, respectively.
Moreover, $\model$ outperforms the baselines 
in each model-dataset combination across both tasks, with improvements
of up to $79\%$ over $\msg$ and $75\%$ over $\dbs$, further 
demonstrating the consistency of $\model$ relative to the two baselines.
In Appendix~\ref{app:dbs_divesity_penalty}, we further analyze the 
performance of $\dbs$ under increased diversity penalties, showing 
that even with stronger diversity regularization, it does not close
the performance gap with $\model$.

\subsection{Comparison to the Full IID Sampling}
Next, we compare $\model_{16}$ and $\model_{64}$ to $\msg_{1024}$, which 
generates all $1024$ responses and thus represents an empirical 
upper bound on ASR. This comparison evaluates how closely $\model$ 
approaches the brute-force limit while using substantially fewer 
full-length generations. The results are presented in 
Table~\ref{tab:PDPS_benchResults}.
Although $\model_{16}$ generates only $16$ full-length responses 
(vs. $1024$ for 
$\msg_{1024}$), it achieves more than $80\%$ of $\msg_{1024}$'s ASR 
in $11$ of the $16$ model–dataset combinations.
For $\model_{64}$, the ASR exceeds $97\%$ of the brute-force benchmark 
across all sixteen model-dataset combinations.
These results demonstrate that $\model$ can produce compact response 
sets while maintaining high attack success rates and substantially 
reducing the number of generations.

\begin{table*}[!t]
\centering
\footnotesize
\caption{Comparison of ASR obtained from $\model$ with IID sampling ($\msg$)
and $\text{Diverse Beam Search}$ ($\dbs$) across four models 
(\textbf{$\modelLlamaSev$}, \textbf{$\modelLlamaTh$}, 
\textbf{$\modelQwenSev$}, \textbf{$\modelQwenFourt$}) and four datasets 
(\textbf{$\dsJBB$}, \textbf{$\dsMalI$}, \textbf{$\dsHarmB$}, 
\textbf{$\dsAdvB$}). Subscripts denote the number of full-length 
generations (e.g., $\msg_{16}$ uses 16 full-length response generations). 
The \textcolor{blue}{blue} entries indicate the performance improvement 
of $\model$ (in $\%$) over the corresponding baseline.}
\label{tab:PDPS_mainResults}
\scalebox{0.97}{
\begin{tabular}{c|p{0.3cm}p{0.4cm}p{0.4cm}p{0.55cm}|p{0.4cm}p{0.4cm}p{0.4cm}p{0.55cm}|p{0.4cm}p{0.4cm}p{0.4cm}p{0.55cm}|p{0.4cm}p{0.4cm}p{0.4cm}p{0.55cm}|p{0.5cm} }
\toprule
& \multicolumn{4}{c|}{\textbf{$\modelLlamaSev$}} & \multicolumn{4}{c|}{\textbf{$\modelLlamaTh$}} & \multicolumn{4}{c|}{\textbf{$\modelQwenSev$}} &  \multicolumn{4}{c|}{\textbf{$\modelQwenFourt$}} & 
\textbf{Avg.} \\
\textbf{} & \textbf{$\dsAdvB$} & \textbf{$\dsJBB$} & \textbf{$\dsMalI$} & \textbf{$\dsHarmB$} & \textbf{$\dsAdvB$} & \textbf{$\dsJBB$} & \textbf{$\dsMalI$} & \textbf{$\dsHarmB$} & \textbf{$\dsAdvB$} & \textbf{$\dsJBB$} & \textbf{$\dsMalI$} & \textbf{$\dsHarmB$} & \textbf{$\dsAdvB$} & \textbf{$\dsJBB$} & \textbf{$\dsMalI$} & \textbf{$\dsHarmB$} & \\
\midrule

% $^{\Diamond}\msg_{1024}$ & 0.73 & 0.90 & 0.94 & 0.88 & 1 & 0.99 & 1.00 & 1.00 & 0.98 & 0.99 & 1.00 & 1.00 & 0.99 & 0.97 & 1.00 & 0.99\\

% \midrule
\rowcolor{yellow!40}
$^{\S}\model_{16}$  & 0.77 & 0.90 & 0.89 & 0.87 
                    & 0.74 & 0.89 & 0.76 & 0.89 
                    & 0.71 & 0.88 & 0.89 & 0.94	
                    & 0.74 & 0.78 & 0.77 & 0.84 & \bf{0.83}\\

\cline{2-18}

$^\clubsuit$\msg$_{16}$ & 0.09 & 0.21  & 0.23 & 0.31 & 0.29 & 0.56 & 0.45 & 0.63 & 0.49	& 0.63 & 0.82 & 0.75 & 0.26 & 0.41 & 0.58 & 0.53 & \bf{0.45}\\
$\triangle^{\S\sim\clubsuit}$ (in \%) & 
\textcolor{blue}{68$\uparrow$} & \textcolor{blue}{69$\uparrow$} & \textcolor{blue}{66$\uparrow$} & \textcolor{blue}{56$\uparrow$} & 
\textcolor{blue}{35$\uparrow$} & \textcolor{blue}{43$\uparrow$} & \textcolor{blue}{37$\uparrow$} & \textcolor{blue}{24$\uparrow$} & 
\textcolor{blue}{22$\uparrow$} & \textcolor{blue}{25$\uparrow$} & \textcolor{blue}{07$\uparrow$} & \textcolor{blue}{19$\uparrow$} & 
\textcolor{blue}{48$\uparrow$} & \textcolor{blue}{37$\uparrow$} & \textcolor{blue}{19$\uparrow$} & \textcolor{blue}{31$\uparrow$} &
\textcolor{blue}{\bf{38}$\mathbf{\uparrow}$} \\

\cline{2-18}

$^{\dagger}\dbs_{16}$ & 0.07 & 0.20 & 0.18 & 0.34 & 0.21 & 0.27 & 0.24 & 0.48 & 0.57 & 0.77 & 0.85 & 0.81 & 0.33 & 0.47 & 0.61 & 0.52 & \bf{0.43}\\
$\triangle^{\S\sim\dagger}$ (in \%)& 
\textcolor{blue}{70$\uparrow$} & \textcolor{blue}{70$\uparrow$} & \textcolor{blue}{71$\uparrow$} & \textcolor{blue}{53$\uparrow$} & 
\textcolor{blue}{53$\uparrow$} & \textcolor{blue}{62$\uparrow$} & \textcolor{blue}{52$\uparrow$} & \textcolor{blue}{41$\uparrow$} & 
\textcolor{blue}{14$\uparrow$} & \textcolor{blue}{11$\uparrow$} & \textcolor{blue}{4$\uparrow$} & \textcolor{blue}{13$\uparrow$} & 
\textcolor{blue}{41$\uparrow$} & \textcolor{blue}{31$\uparrow$} & \textcolor{blue}{16$\uparrow$} & \textcolor{blue}{32$\uparrow$} &
\textcolor{blue}{\bf{40}$\mathbf{\uparrow}$} \\

\midrule
\rowcolor{yellow!40}

$^{\P}\model_{64}$ & 0.81 & 0.93 & 0.94 & 0.94 
                   & 0.97 & 0.98 & 0.99 & 1.00
                   & 0.97 & 0.97 & 1.00 & 0.99 
                   & 0.98 & 0.99 & 0.98 & 0.97 & \bf{0.96} \\

\cline{2-18}

$^\spadesuit\msg_{64}$ & 0.18 & 0.48 & 0.55	& 0.50 & 0.67 & 0.81 & 0.75 & 0.86 & 0.75  & 0.87 & 0.95 & 0.90 & 0.60 & 0.73 & 0.84 & 0.77 & \bf{0.70}\\
$\triangle^{\P\sim\spadesuit}$ (in \%) & 
\textcolor{blue}{63$\uparrow$} & \textcolor{blue}{45$\uparrow$} & \textcolor{blue}{39$\uparrow$} & \textcolor{blue}{44$\uparrow$} & 
\textcolor{blue}{30$\uparrow$} & \textcolor{blue}{17$\uparrow$} & \textcolor{blue}{24$\uparrow$} & \textcolor{blue}{14$\uparrow$} & 
\textcolor{blue}{22$\uparrow$} & \textcolor{blue}{10$\uparrow$} & \textcolor{blue}{05$\uparrow$} & \textcolor{blue}{09$\uparrow$} & 
\textcolor{blue}{38$\uparrow$} & \textcolor{blue}{26$\uparrow$} & \textcolor{blue}{14$\uparrow$} & \textcolor{blue}{20$\uparrow$} & 
\textcolor{blue}{\bf{26}$\mathbf{\uparrow}$} \\

\cline{2-18}

$^{\ddagger}\dbs_{64}$& 0.22 & 0.35 & 0.43 & 0.54 & 0.22 & 0.35 & 0.37 & 0.56 & 0.88 & 0.92 & 0.96 & 0.99 &  0.69 & 0.72  & 0.89  & 0.74 & \bf{0.61}\\
$\triangle^{\P\sim\ddagger}$ (in \%) & 
\textcolor{blue}{55$\uparrow$} & \textcolor{blue}{56$\uparrow$} & \textcolor{blue}{52$\uparrow$} & \textcolor{blue}{38$\uparrow$} & 
\textcolor{blue}{75$\uparrow$} & \textcolor{blue}{63$\uparrow$} & \textcolor{blue}{62$\uparrow$} & \textcolor{blue}{41$\uparrow$} & 
\textcolor{blue}{09$\uparrow$} & \textcolor{blue}{05$\uparrow$} & \textcolor{blue}{04$\uparrow$} & \textcolor{black}{00} & 
\textcolor{blue}{29$\uparrow$} & \textcolor{blue}{27$\uparrow$} & \textcolor{blue}{09$\uparrow$} & \textcolor{blue}{23$\uparrow$} &
\textcolor{blue}{\bf{35}$\mathbf{\uparrow}$} \\

\bottomrule

\end{tabular}
}

\end{table*}

\begin{table*}[!t]
\centering
\footnotesize\
\caption{ASR of $\model$ compared to the brute-force benchmark $\msg_{1024}$ 
across the four models (\textbf{$\modelLlamaSev$}, \textbf{$\modelLlamaTh$}, 
\textbf{$\modelQwenSev$}, \textbf{$\modelQwenFourt$}) and the four datasets 
(\textbf{$\dsJBB$}, \textbf{$\dsMalI$}, \textbf{$\dsHarmB$}, 
\textbf{$\dsAdvB$}). Subscripts denote the number of full-length generations
(e.g., $\model_{16}$ uses $16$ full-length sequences constructed from $1024$ 
partial generations). The row labeled $\rho^{\S\sim \Diamond}$ (resp. 
$\rho^{\P\sim \Diamond}$) report the ASR achieved by $\model_{16}$ 
(resp. $\model_{64}$) as a fraction of the ASR achieved by $\msg_{1024}$. The \textcolor{blue}{blue} entries in these rows indicate that $\model$ attains at least $0.8$ times
the ASR of $\msg_{1024}$, whereas \textcolor{red}{red} entries indicate failure to reach this 
threshold.
}
\label{tab:PDPS_benchResults}
\begin{tabular}{c|p{0.45cm}p{0.45cm}p{0.45cm}p{0.55cm}|p{0.45cm}p{0.45cm}p{0.45cm}p{0.55cm}|p{0.45cm}p{0.45cm}p{0.45cm}p{0.55cm}|p{0.45cm}p{0.45cm}p{0.45cm}p{0.55cm} }
\toprule
& \multicolumn{4}{c|}{\textbf{$\modelLlamaSev$}} & \multicolumn{4}{c|}{\textbf{$\modelLlamaTh$}} & \multicolumn{4}{c|}{\textbf{$\modelQwenSev$}} &  \multicolumn{4}{c}{\textbf{$\modelQwenFourt$}} \\
\textbf{} & \textbf{$\dsAdvB$} & \textbf{$\dsJBB$} & \textbf{$\dsMalI$} & \textbf{$\dsHarmB$} & \textbf{$\dsAdvB$} & \textbf{$\dsJBB$} & \textbf{$\dsMalI$} & \textbf{$\dsHarmB$} & \textbf{$\dsAdvB$} & \textbf{$\dsJBB$} & \textbf{$\dsMalI$} & \textbf{$\dsHarmB$} & \textbf{$\dsAdvB$} & \textbf{$\dsJBB$} & \textbf{$\dsMalI$} & \textbf{$\dsHarmB$}  \\
\midrule

$^{\Diamond}\msg_{1024}$ & 0.73 & 0.88 & 0.95 & 0.97 & 1.00 & 0.99 & 1.00 & 1.00 & 0.98 & 0.99 & 1.00 & 1.00 & 0.99 & 0.97 & 1.00 & 0.99\\

\midrule

%\rowcolor{yellow!40}
$^{\S}\model_{16}$  & 0.77 & 0.90 & 0.89 & 0.87 & 0.74 & 0.89 & 0.76 & 0.89	& 0.71 & 0.88 & 0.89 & 0.94	& 0.74 & 0.78 & 0.77 & 0.84\\
$\rho^{\S\sim \Diamond}$ & 
\textcolor{blue}{$\ge$1.0} & \textcolor{blue}{$\ge$1.0} & \textcolor{blue}{0.94} & \textcolor{blue}{0.90} & 
\textcolor{red}{0.74} & \textcolor{blue}{0.90} & \textcolor{red}{0.76} & \textcolor{blue}{0.89} & 
\textcolor{red}{0.72} & \textcolor{blue}{0.89} & \textcolor{blue}{0.89} & \textcolor{blue}{0.94} & 
\textcolor{red}{0.75} & \textcolor{blue}{0.80} & \textcolor{red}{0.77} & \textcolor{blue}{0.85} \\

\cline{2-17}

%\rowcolor{yellow!40}

$^{\P}\model_{64}$ & 0.81 & 0.93 & 0.94 & 0.94 & 0.97 & 0.98 & 0.99 & 1.0 & 0.97 & 0.97 & 1.00 & 0.99 & 0.98 & 0.99 & 0.98 & 0.97\\
$\rho^{\P\sim \Diamond}$ & 
\textcolor{blue}{$\ge$1.0} & \textcolor{blue}{$\ge$1.0} & \textcolor{blue}{0.99} & \textcolor{blue}{0.97} & 
\textcolor{blue}{0.97} & \textcolor{blue}{0.99} & \textcolor{blue}{0.99} & \textcolor{blue}{$\ge$1.0} & 
\textcolor{blue}{0.99} & \textcolor{blue}{0.98} & \textcolor{blue}{1.0} & \textcolor{blue}{0.99} & 
\textcolor{blue}{0.99} & \textcolor{blue}{$\ge$1.0} & \textcolor{blue}{0.98} & \textcolor{blue}{0.98} \\

\bottomrule

\end{tabular}

\end{table*}

\subsection{Failure Mode Coverage and Diversity Analysis of Unsafe Responses}
\label{sec:div_analysis_unsafe_resp}
In this section, we examine the coverage and diversity of failure modes 
identified by $\model$ in comparison to the baseline methods. While the overall ASR valued provide a coarse measure of effectiveness, they do not reveal
whether a method uncovers a broad spectrum of vulnerabilities or repeatedly
triggers the same failure pattern.
In particular, a method may fail to generate any unsafe response for 
certain queries, thereby appearing weaker in terms of ASR. However, even
when a method succeeds in producing unsafe outputs, an important question 
remains: do these outputs correspond to diverse and distinct failure
modes, or are they concentrated around a limited set of similar behaviors?

\begin{figure}[t!]
    \centering
    \begin{tikzpicture}

\begin{groupplot}[
    group style={
        group size=2 by 2,
        horizontal sep=1.2cm,
        vertical sep=1.2cm,
        xticklabels at=edge bottom
    },
    width=0.42\linewidth,
    height=3.5cm,
    ymin=-5,
    ymax=5,
    symbolic x coords={$\dsAdvB$, $\dsJBB$, $\dsHarmB$, $\dsMalI$},
    enlarge x limits=0.25,
    clip=false,
    extra y ticks={0},
    extra y tick style={grid=major},
]

% ================== Plot 1 ==================
\nextgroupplot[
    title={$\modelLlamaSev$},
    ybar,
    bar width=6pt,
    ylabel={Mean Toxicity Diff.},
    ylabel style={font=\small},
]
% ---- MSG (left bars) ----
\addplot[
    bar shift=-4pt,
    fill=blue
] coordinates {
    ($\dsAdvB$,3.42)
    ($\dsJBB$,1.42)
    ($\dsHarmB$,3.02)
    ($\dsMalI$,4.34)
};

% ---- DBS (right bars) ----
\addplot[
bar shift=4pt,
    fill=blue!20,
    pattern=north east lines,
    pattern color=blue
] coordinates {
    ($\dsAdvB$,3.32)
    ($\dsJBB$,2.22)
    ($\dsHarmB$,1.36)
    ($\dsMalI$,4.20)
};

% ================== Plot 2 ==================
\nextgroupplot[
    title={$\modelLlamaTh$},
    ybar,
    bar width=6pt
]
% ---- MSG (left bars) ----
\addplot[
    bar shift=-4pt,
    fill=red
] coordinates {
    ($\dsHarmB$,-1.68)
};

\addplot[
    bar shift=-4pt,
    fill=blue
] coordinates {
    ($\dsAdvB$,2.25)
    ($\dsJBB$,3.37)
    ($\dsMalI$,2.69)
};

% ---- DBS (right bars) ----
\addplot[
bar shift=4pt,
    fill=red!20,
    pattern=north east lines,
    pattern color=red
] coordinates {
    ($\dsHarmB$,-2.33)
};

\addplot[
bar shift=4pt,
    fill=blue!20,
    pattern=north east lines,
    pattern color=blue
] coordinates {
    ($\dsAdvB$,3.57)
    ($\dsJBB$,2.97)
    ($\dsMalI$,3.75)
};

% ================== Plot 3 ==================
\nextgroupplot[
    title={$\modelQwenSev$},
    ybar,
    bar width=6pt,
    ylabel={Mean Toxicity Diff.},
    ylabel style={font=\small},
]
% ---- MSG (left bars) ----
\addplot[
    bar shift=-4pt,
    fill=red
] coordinates {
    ($\dsHarmB$,-2.09)
    ($\dsMalI$,-3.84)
};

\addplot[
    bar shift=-4pt,
    fill=blue
] coordinates {
    ($\dsAdvB$,1.54)
    ($\dsJBB$,1.29)
};

% ---- DBS (right bars) ----
\addplot[
bar shift=4pt,
    fill=red!20,
    pattern=north east lines,
    pattern color=red
] coordinates {
    ($\dsAdvB$,-1.71)
    ($\dsJBB$,-2.35)
    ($\dsHarmB$,-4.42)
    ($\dsMalI$,-4.50)
};

% ================== Plot 4 ==================
\nextgroupplot[
    title={$\modelQwenFourt$},
    ybar,
    bar width=6pt
]
% ---- MSG (left bars) ----
\addplot[
    bar shift=-4pt,
    fill=blue
] coordinates {
    ($\dsAdvB$,2.97)
    ($\dsJBB$,1.56)
    ($\dsHarmB$,0.21)
    ($\dsMalI$,2.48)
};

% ---- DBS (right bars) ----
\addplot[
bar shift=4pt,
    fill=red!20,
    pattern=north east lines,
    pattern color=red
] coordinates {
    ($\dsHarmB$,-1.93)
    ($\dsMalI$,-1.73)
};

\addplot[
bar shift=4pt,
    fill=blue!20,
    pattern=north east lines,
    pattern color=blue
] coordinates {
    ($\dsAdvB$,1.99)
    ($\dsJBB$,0.36)
};

\end{groupplot}

% ---------- Common Legend ----------
\begin{scope}[shift={(0,-4.8cm)}]
\matrix[
    matrix of nodes,
    anchor=north,
    nodes={
        anchor=west,
        inner sep=1pt,
        outer sep=0pt
    },
    column sep=3pt
] at (current bounding box.south) {

    \tikz\draw[fill=blue] (0,0) rectangle (0.25,0.25); &
    $\msg_{64}$ &
    \hspace{12pt} % ← space between legend groups
    % \tikz\draw[fill=red] (0,0) rectangle (0.25,0.25); &
    % $\msg_{64}$ - &
    % \hspace{12pt} % ← space between legend groups
    \tikz\draw[
        fill=blue!20,
        pattern=north east lines,
        pattern color=blue
    ] (0,0) rectangle (0.25,0.25); &
    $\dbs_{64}$ \\
    % \hspace{12pt} % ← space between legend groups
    % \tikz\draw[
    %     fill=red!20,
    %     pattern=north east lines,
    %     pattern color=red
    % ] (0,0) rectangle (0.25,0.25); &
    % $\dbs_{64}$ - \\
};
\end{scope}

\end{tikzpicture}
    \caption{Bar plot of the mean toxicity difference, defined as the difference
    between the average number of unsafe responses returned by $\model$ and that
    returned by a baseline method, across various model–dataset combinations. The
    average is computed over only those queries for which the respective method 
    achieves a successful attack. Blue bars indicate a positive difference, while
    red bars indicate a negative difference.}
    \label{fig:mean_toxicity_diff}
\end{figure}
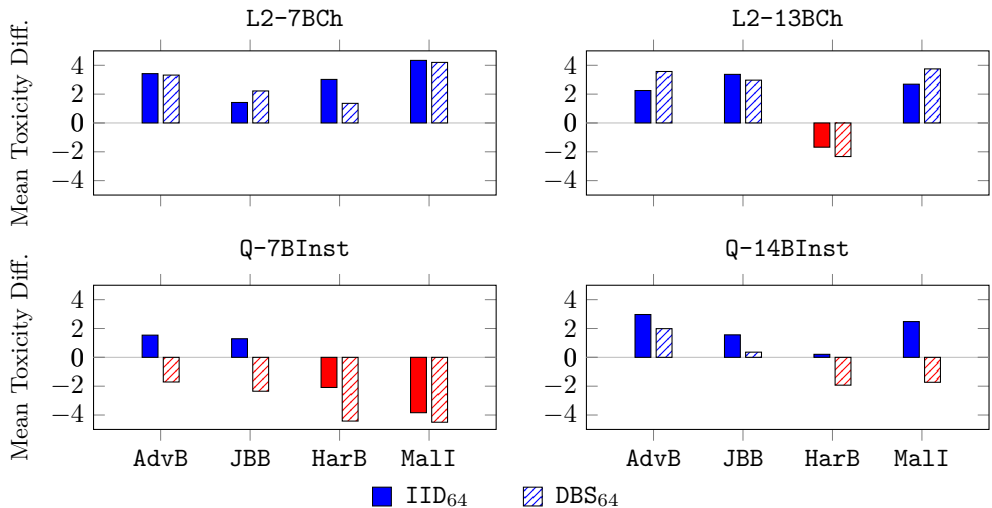

\begin{table}
    \centering
    \footnotesize
    \caption{Comparison of the average diversity of unsafe responses generated
    by $\model$ and two baseline methods for the $\modelQwenSev$ model on the
    four datasets. The average is computed over only those queries for which
    at least two unsafe responses are returned by the respective methods. 
    Diversity metrics: Dist-$n$ = Distinct-$n$~\citep{li2016diversity}; 
    SB-$n$ = Self-BLEU-$n$~\citep{zhu2018texygen}; 
    Uni-Ent = Unigram Entropy~\citep{csaky2019improving};
    Cos-Dist = Cosine Distance~\citep{jiang2011similarity}; 
    BERT-Div = BERTScore Diversity~\citep{zhang2019bertscore}. 
    For metrics marked with $\uparrow$, higher values indicate greater 
    diversity, while $\downarrow$ indicates the opposite.}
    \label{tab:unsafe_diversity_qwen7b}
    \begin{tabular}{c|c|ccccccccc}
    \toprule
     Dataset & Sampler
          & Dist-1$\uparrow$ 
          & Dist-2$\uparrow$ 
          & SB-1$\downarrow$ 
          & SB-2$\downarrow$ 
          & SB-3$\downarrow$ 
          & SB-4$\downarrow$ 
          & Uni-Ent$\uparrow$ 
          & Cos-Dist$\uparrow$ 
          & BERT-Div$\uparrow$ \\
    \midrule
    & $\msg_{64}$ & 0.27 & 0.68 & 0.62 & 0.42 & 0.28 & 0.19 & 5.46 & 0.31 & 0.29 \\
    $\dsAdvB$ & $\dbs_{64}$ & 0.17 & 0.48 & 0.74 & 0.58 & 0.47 & 0.39 & 5.31 & 0.24 & 0.25 \\
    &{\bf$\model_{64}$} & {\bf 0.32} & {\bf 0.77} & {\bf 0.59} & {\bf 0.34} & {\bf 0.20} & {\bf 0.12} & {\bf 5.81} & {\bf 0.47} & {\bf 0.34} \\
    
    \midrule
    
    & $\msg_{64}$ & 0.24 & 0.65 & 0.69 & 0.48 & 0.33 & 0.23 & 5.57 & 0.26 & 0.29 \\
    $\dsJBB$ & $\dbs_{64}$ & 0.15 & 0.45 & 0.80 & 0.65 & 0.53 & 0.44 & 5.39 & 0.18 & 0.25 \\
    &{\bf$\model_{64}$} & {\bf 0.30} & {\bf 0.76} & {\bf 0.65} & {\bf 0.39} & {\bf 0.23} & {\bf 0.14} & {\bf 6.00} & {\bf 0.40} & {\bf 0.34} \\

    \midrule

    & $\msg_{64}$ & 0.20 & 0.58 & 0.76 & 0.57 & 0.41 & 0.30 & 5.66 & 0.22 & 0.28 \\
    $\dsHarmB$ & $\dbs_{64}$ & 0.12 & 0.38 & 0.84 & 0.71 & 0.60 & 0.51 & 5.39 & 0.17 & 0.24 \\
    &{\bf$\model_{64}$} & {\bf 0.26} & {\bf 0.71} & {\bf 0.71} & {\bf 0.46} & {\bf 0.29} & {\bf 0.19} & {\bf 6.06} & {\bf 0.36} & {\bf 0.34} \\

    \midrule

    & $\msg_{64}$ & 0.19 & 0.58 & 0.74 & 0.53 & 0.38 & 0.26 & 5.58 & 0.32 & 0.29 \\
    $\dsMalI$ & $\dbs_{64}$ & 0.13 & 0.39 & 0.81 & 0.66 & 0.54 & 0.45 & 5.32 & 0.22 & 0.25 \\
    &{\bf$\model_{64}$} & {\bf 0.30} & {\bf 0.77} & {\bf 0.60} & {\bf 0.34} & {\bf 0.19} & {\bf 0.11} & {\bf 5.99} & {\bf 0.55} & {\bf 0.37} \\
    
    \bottomrule
    \end{tabular}

\end{table}

To move beyond ASR, we analyze both the number and diversity of unsafe 
responses per successful query. We first compare the average number of
unsafe responses per query, conditioning on queries with at least one 
unsafe output. This isolates a method’s ability to uncover multiple 
failure modes independent of overall ASR. We then evaluate the average 
diversity of unsafe responses to ensure additional samples provide 
distinct insights rather than redundant variations.

\paragraph{Analysis of Mean Toxicity Difference.} 
Figure~\ref{fig:mean_toxicity_diff} presents the mean toxicity difference, 
defined as the difference between the average number of unsafe responses 
identified by $\model$ and those identified by the baseline methods, 
across various 
model–dataset combinations. $\model$ consistently 
detects a higher number of unsafe responses than $\msg$ across nearly all 
combinations. While the comparison with $\dbs$ yields mixed results for 
the Qwen models, $\model$ consistently identifies more unsafe responses 
for the Llama models.

\paragraph{Analysis of Diversity.}
We further assess each method’s ability to uncover distinct failure 
modes by measuring the average diversity of unsafe responses, computed
over queries with at least two unsafe outputs. 
Table~\ref{tab:unsafe_diversity_qwen7b} reports nine diversity metrics
for $\modelQwenSev$ across four datasets. $\model$ outperforms both 
baselines across all metrics and datasets, demonstrating broader 
failure coverage. Results for the remaining models 
(Appendix~\ref{app:diversity_analysis}) indicate that $\model$ 
consistently outperforms $\dbs$ by a wide margin across all metrics 
and outperforms $\msg$ across most metrics.

Overall, these findings indicate that $\model$ more effectively 
uncovers diverse failure modes than competing methods. 
Appendix~\ref{app:qualitative_diversity_analysis} provides qualitative 
examples, showing that $\dbs$ tends to produce minor surface 
variations, whereas $\model$ generates more semantically distinct 
responses.

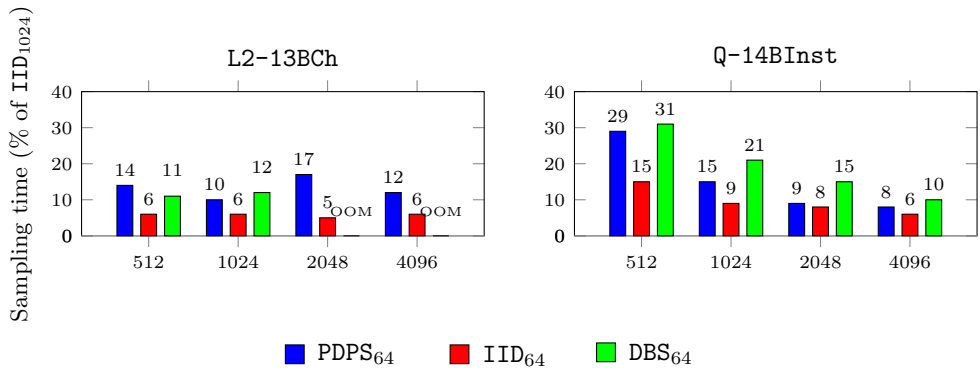
\begin{figure}[t]
\centering
\begin{tikzpicture}

\begin{groupplot}[
    group style={
        group size=2 by 1,
        horizontal sep=1.2cm,
        vertical sep=1.2cm,
        xticklabels at=edge bottom
    },
    width=0.42\linewidth,
    height=3.5cm,
    ymin=0,
    ymax=40,
    symbolic x coords={$512$, $1024$, $2048$, $4096$},
    enlarge x limits=0.25,
    clip=false,
    extra y ticks={0},
    extra y tick style={grid=major},
    tick label style={font=\scriptsize}
]

% ================== Plot 1 ==================
\nextgroupplot[
    title={$\modelLlamaTh$},
    ybar,
    bar width=6pt,
    ylabel={Sampling time (\% of $\msg_{1024}$)},
    ylabel style={font=\small},
]
% ---- PDPS (left bars) ----
\addplot[
    bar shift=-9pt,
    fill=blue,
    nodes near coords,
    point meta=y,
    nodes near coords style={
        font=\scriptsize,
        /pgf/number format/fixed,
        /pgf/number format/precision=2
    },
] coordinates {
    ($512$,14)
    ($1024$,10)
    ($2048$,17)
    ($4096$,12)
};

% ---- IID (middle bars) ----
\addplot[
    bar shift=0pt,
    fill=red,
    nodes near coords,
    point meta=y,
    nodes near coords style={
        font=\scriptsize,
        /pgf/number format/fixed,
        /pgf/number format/precision=2
    },
] coordinates {
    ($512$,06)
    ($1024$,06)
    ($2048$,05)
    ($4096$,06)
};

% ---- DBS (right bars) ----
\addplot[
    bar shift=9pt,
    fill=green,
    point meta=y,
    nodes near coords={
        \pgfmathfloatparsenumber{\pgfplotspointmeta}%
        \pgfmathfloattofixed{\pgfmathresult}%
        \let\temp=\pgfmathresult
        \ifdim\temp pt=0pt
            {\tiny OOM}
        \else
            \pgfmathprintnumber[fixed,precision=2]{\temp}
        \fi
    },
    every node near coord/.append style={
        font=\scriptsize,
        yshift=4pt
    },
] coordinates {
    ($512$,11)
    ($1024$,12)
    ($2048$,0) % dummy
    ($4096$,0) % dummy
};

% ================== Plot 2 ==================
\nextgroupplot[
    title={$\modelQwenFourt$},
    ybar,
    bar width=6pt
]

% ---- PDPS (left bars) ----
\addplot[
    bar shift=-9pt,
    fill=blue,
    nodes near coords,
    point meta=y,
    nodes near coords style={
        font=\scriptsize,
        /pgf/number format/fixed,
        /pgf/number format/precision=2
    },
] coordinates {
    ($512$,29)
    ($1024$,15)
    ($2048$,09)
    ($4096$,08)
};

% ---- IID (middle bars) ----
\addplot[
    bar shift=0pt,
    fill=red,
    nodes near coords,
    point meta=y,
    nodes near coords style={
        font=\scriptsize,
        /pgf/number format/fixed,
        /pgf/number format/precision=2
    },
] coordinates {
    ($512$,15)
    ($1024$,09)
    ($2048$,08)
    ($4096$,06)
};

% ---- DBS (right bars) ----
\addplot[
    bar shift=9pt,
    fill=green,
    nodes near coords,
    point meta=y,
    nodes near coords style={
        font=\scriptsize,
        /pgf/number format/fixed,
        /pgf/number format/precision=2
    },
] coordinates {
    ($512$,31)
    ($1024$,21)
    ($2048$,15)
    ($4096$,10)
};
\end{groupplot}

% ---------- Common Legend ----------
\begin{scope}[shift={(0,-4.8cm)}]
\matrix[
    matrix of nodes,
    anchor=north,
    nodes={
        anchor=west,
        inner sep=1pt,
        outer sep=0pt
    },
    column sep=3pt
] at (current bounding box.south) {

    \tikz\draw[fill=blue] (0,0) rectangle (0.25,0.25); &
    $\model_{64}$ &
    \hspace{12pt} % ← space between legend groups
    \tikz\draw[fill=red] (0,0) rectangle (0.25,0.25); &
    $\msg_{64}$
    \hspace{12pt} % ← space between legend groups
    \tikz\draw[fill=green] (0,0) rectangle (0.25,0.25); &
    $\dbs_{64}$ \\
};
\end{scope}

\end{tikzpicture}
\caption{Sampling time (as a percentage of brute-force $\msg_{1024}$) for
generating 64 responses on $\modelLlamaTh$ and $\modelQwenFourt$
across token lengths $512$--$4096$. Bars labeled \emph{OOM}
indicate out-of-memory failures.}
\label{fig:execution_time}
\end{figure}

{
%\color{blue}

\subsection{Safety Hardening via Adversarial Fine-Tuning}
\label{sec:hardening}

The preceding analyses establish that $\model$ uncovers a broader and
more semantically diverse set of failure modes than $\msg$ and
$\dbs$. A natural question is whether this increased diversity translates
into more effective safety hardening when the generated toxic responses
are used as negative training signals. To investigate this question, we
conduct a fine-tuning experiment in which a base model is further
safety-tuned using toxic responses generated by each method as negative
examples and refusal responses as positive examples within an RLHF
pipeline. We then evaluate the resulting models on their resistance to
jailbreak attacks.

\paragraph{Setup.} For each method, $\model$, $\msg$, and $\dbs$, we
collect the toxic responses using a $64$-response generation task and use
them as negative samples for LoRA fine-tuning with 
GRPO~\citep{shao2024deepseekmath}, paired with the 
model's corresponding refusal responses as positive samples. The 
resulting finetuned models are evaluated using $\msg_{64}$. For each 
method, the base model was fine-tuned for 25 epochs using the AdamW 
optimizer.

\begin{table}[h]
\centering
\caption{ASR of the base model and models fine-tuned using negative 
samples generated by $\msg_{64}$, $\dbs_{64}$, and $\model_{64}$. 
Lower ASR indicates stronger safety hardening.}
\label{tb:hardening}
\begin{tabular}{lc}
\toprule
\textbf{Method} & \textbf{ASR} \\
\midrule
Base Model (no fine-tuning) & 0.75 \\
\midrule
Fine-tuned w/ $\msg_{64}$ & 0.36 \\
Fine-tuned w/ $\dbs_{64}$ & 0.41 \\
Fine-tuned w/ $\model_{64}$ & \textbf{0.24} \\
\bottomrule
\end{tabular}
\end{table}

\paragraph{Results.} Table~\ref{tb:hardening} reports the ASR of the
base model and the three fine-tuned variants. The base model achieves an
ASR of $0.75$, indicating that substantial vulnerability remains even
after standard safety tuning. Fine-tuning with negative samples
generated by $\msg_{64}$ and $\dbs_{64}$ reduces the ASR to $0.36$ and
$0.41$, respectively, confirming that adversarial fine-tuning improves
robustness against jailbreak attacks. However, fine-tuning with
$\model_{64}$-generated outputs achieves the lowest ASR of $24\%$,
corresponding to a relative reduction of $33\%$ compared to
$\msg_{64}$ and $41\%$ compared to $\dbs_{64}$. We attribute this
improvement to the broader coverage of distinct failure modes provided
by $\model$, which enables the safety-tuning process to mitigate a
wider range of vulnerabilities.

}

\subsection{Computational Efficiency Analysis}
In this section, we analyze the computational gain of $\model$
for the 64-response generation task, for which $\model$ already
achieves an ASR comparable to the brute-force $\msg_{1024}$ upper
bound (see Table~\ref{tab:PDPS_benchResults}).
For the 64-response setting, $\msg_{64}$, which performs IID sampling
of only 64 responses per query, provides a lower bound on computational
cost. Ideally, $\msg_{64}$ requires approximately $1/16 = 0.06$ (i.e., 
$6\%$) of the time required by $\msg_{1024}$. Therefore, we evaluate 
how closely the runtime of $\model_{64}$ approaches this lower bound. 
We also compare against $\dbs_{64}$ as a representative 
diversity-inducing baseline.
Figure~\ref{fig:execution_time} reports the sampling time of 
$\model_{64}$, $\msg_{64}$, and $\dbs_{64}$ as a percentage 
of the brute-force upper bound $\msg_{1024}$ for two models, 
$\modelLlamaTh$ and $\modelQwenFourt$, across token generation 
lengths ranging from $512$ to $4096$.

\paragraph{Results on $\modelQwenFourt$.}
Figure~\ref{fig:execution_time} (right), showing the results for $\modelQwenFourt$, indicates that for shorter generation lengths
(e.g., $512$ tokens), the sampling time of $\msg_{64}$ is noticeably
higher than the ideal $6\%$ lower bound. As the generation length 
increases, the runtime approaches this limit.
We attribute this deviation at shorter lengths to suboptimal GPU 
utilization caused by the smaller batch size of $\msg_{64}$ compared
to $\msg_{1024}$. As the generation length increases, GPU utilization
improves, enabling $\msg_{64}$ to approach the theoretical $6\%$ 
bound.
A similar trend is observed for $\model_{64}$, which requires as much
as $29\%$ of the $\msg_{1024}$ runtime at $512$ tokens. However, its 
relative overhead decreases with longer generation lengths, reaching
approximately $8\%$ at $4096$ tokens. Compared to $\msg_{64}$, 
$\model_{64}$ incurs higher overhead at shorter lengths, 
but this gap narrows as generation length increases.
This behavior is expected because $\model_{64}$ initially generates 
$1024$ short partial responses. When the final generation length is 
small, this initial cost becomes a significant part of the
total runtime; however, it becomes negligible for longer sequences.
The computational trend of $\dbs_{64}$ is similar, although its runtime
remains slightly higher than that of $\model_{64}$.

% \footnotetext{The implementations used to measure sampling time 
% process each query independently, i.e., each query constitutes 
% a single batch for response generation. No batching across 
% multiple queries is performed.}

\paragraph{Results on $\modelLlamaTh$.}
Figure~\ref{fig:execution_time} (left) shows the results for 
$\modelLlamaTh$. In contrast to $\modelQwenFourt$, $\msg_{64}$ achieves 
the $6\%$ lower bound across all generation lengths. Investigation 
reveals significantly better GPU utilization for this model, allowing 
near-optimal efficiency even at shorter lengths.
For $\model_{64}$, the sampling time ranges between $10\%$ and $17\%$ 
across token lengths, with no clear trend toward the $6\%$ lower bound 
as the token generation length increases. Further analysis suggests that 
this behavior arises from uneven GPU utilization across different 
generation stages of $\model$.
Despite this suboptimality, $\model_{64}$ requires on average only 
$13\%$ of the brute-force runtime, approximately twice the theoretical 
$6\%$ lower bound, while achieving performance comparable to $\msg_{1024}$ 
(Table~\ref{tab:PDPS_benchResults}). Improving the implementation of 
$\model$ to enhance execution efficiency remains an avenue for future 
work.
$\dbs_{64}$ results in out-of-memory (OOM) errors on 
an A100 GPU for generation lengths exceeding $2048$. For shorter
lengths, its runtime is comparable to that of $\model_{64}$.

Overall, the results demonstrate that $\model$ achieves performance
comparable to the brute-force $\msg_{1024}$ upper bound while reducing
the sampling time to $8\%-29\%$ of that required by $\msg_{1024}$.

\subsection{Hyperparameter Sensitivity}
\label{sec:hyperparameter_sensitivity}
In Figure~\ref{fig:hyperparameter_tuning}, we present the sensitivity of
$\model$'s ASR to three hyperparameters: (a) the nucleus sampling
probability $p$, (b) the temperature $\tau$, and (c) $\lambda$, which
controls the quality–diversity trade-off. The results indicate that
increasing these hyperparameters generally improves ASR up to a certain
point, primarily due to the increased diversity of generated samples.
However, excessive diversity can reduce ASR by producing incoherent
responses. A more detailed discussion is provided in
Appendix~\ref{app:hyperparameter_sensitivity}.

{
%\color{red}

\subsection{Is Input-Space Search a Substitute for Output-Space 
Exploration?}
\label{subsec:input_vs_output}

The fine-tuning results above underscore the importance of
diverse failure mode coverage. This raises a complementary question:
can input-space prompt optimization methods, which search for
adversarial prompts rather than diverse outputs, provide equivalent
or superior coverage? To investigate this, we evaluate the 
output-space search methods against five prominent input-space 
adversarial attack frameworks (refer to 
Section~\ref{subsec:baselines} for details of these methods) 
on the $\modelQwenSev$ model and the $\dsAdvB$ dataset.

Under their standard deployment settings, where optimization is
paired with single-response generation for \textbf{GCG},
\textbf{PiF}, and \textbf{ASETF}, and iterative prompt refinement
with single-response evaluation for \textbf{PAIR} and
\textbf{TAP}, the resulting ASRs are: \textbf{GCG} ($0.48$),
\textbf{PiF} ($0.22$), \textbf{ASETF} ($0.15$), \textbf{PAIR}
($0.29$), and \textbf{TAP} ($0.26$). These results fall well below
all output-space search methods, even at only $16$ generations (ASR
of $\model_{16}$, $\msg_{16}$, and $\dbs_{16}$ are $0.71$, $0.49$,
and $0.57$, respectively), despite the input-space methods incurring
substantially greater execution time (see Figure~\ref{fig:input_vs_output}). 
Beyond their lower ASR, these
methods are inherently limited in their ability to uncover a diverse
range of failure modes through multiple trials, given their higher
per-trial computational cost.

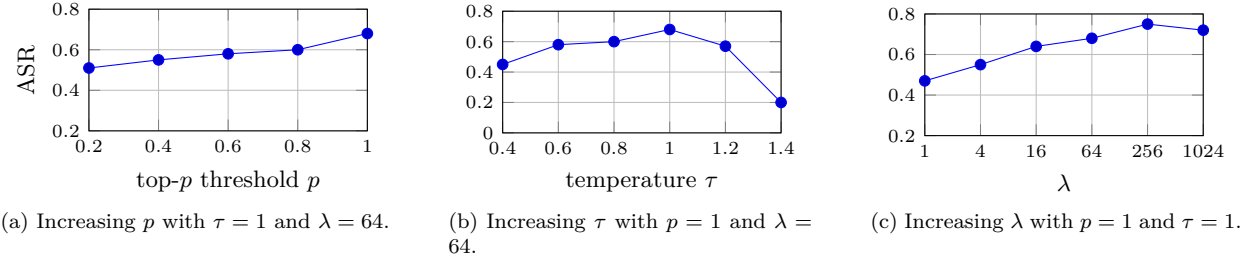
\begin{figure*}[t]
    \centering
    \subfloat[Increasing $p$ with $\tau=1$ and $\lambda=64$.\label{fig:tuning_p}]{
    \begin{tikzpicture}
        \begin{axis}[
            width=0.34\linewidth,
            height=3.2cm,
            ylabel={ASR},
            ymin=0.2,
            ymax=0.8,
            xtick={0.2, 0.4, 0.6, 0.8, 1.0},
            xmin=0.2,
            xmax=1.0,
            xlabel={top-$p$ threshold $p$},
            grid=both,
            legend pos=south east,
            xlabel style={font=\small},
            ylabel style={font=\small},
            tick label style={font=\scriptsize},
            legend style={font=\scriptsize,draw=none,fill=none},
        ]
        % Example curves
        \addplot coordinates { (0.2,0.51)(0.4,0.55)(0.6,.58)(0.8,0.6)(1.0,0.68) };
        %\addlegendentry{advB}

        \end{axis}
    \end{tikzpicture}
   }
   \hfill
% -------- Subfigure 2 --------
    \subfloat[Increasing $\tau$ with $p=1$ and $\lambda=64$.\label{fig:tuning_tau}]{
    \begin{tikzpicture}
        \begin{axis}[
            width=0.34\linewidth,
            height=3.2cm,
            xlabel={temperature $\tau$},
            xtick={0.4, 0.6, 0.8, 1.0, 1.2, 1.4},
            xmin=0.4,
            xmax=1.4,
            ymin=0,
            ymax=0.8,
            grid=both,
            legend pos=north west,
            xlabel style={font=\small},
            ylabel style={font=\small},
            tick label style={font=\scriptsize},
        ]
        \addplot coordinates {(0.4,0.45) (0.6, 0.58) (0.8,0.6)(1.0,0.68)(1.2,0.57)(1.4,0.20)};
        % \addlegendentry{AdvBench}
        \end{axis}
    \end{tikzpicture}  
    } \hfill
    % -------- Subfigure 3 --------
    \subfloat[Increasing $\lambda$ with $p=1$ and $\tau=1$.\label{fig:tuning_lambda}]{
    \begin{tikzpicture}
        \begin{axis}[
            width=0.34\linewidth,
            height=3.2cm,
            xlabel={$\lambda$},
            xtick={1, 4, 16, 64, 256, 1024},
            xticklabels={1,4,16,64,256,1024},
            xmin=1,
            xmax=1024,
            xmode=log,
            ymin=0.2,
            ymax=0.8,
            grid=both,
            legend pos=north west,
            xlabel style={font=\small},
            ylabel style={font=\small},
            tick label style={font=\scriptsize},
        ]
        \addplot coordinates {(1,0.47) (4, 0.55) (16,0.64) (64,0.68) (256,0.75) (1024,0.72)};
        % \addlegendentry{advB}
        \end{axis}
    \end{tikzpicture}
    }
    
    \caption{ASR under different hyperparameter settings for the $\modelQwenSev$ model on the $\dsAdvBFull$ dataset. (a) ASR for different nucleus sampling probabilities ($p$); (b) ASR for different sampling temperatures ($\tau$); and (c) ASR for different values of $\lambda$, the hyperparameter controlling the quality–diversity threshold.}
    \label{fig:hyperparameter_tuning}
\end{figure*}

These findings indicate that input-space search alone cannot replace 
systematic output-space exploration for two key reasons. First, 
input-space methods typically incur high computational overhead 
or require extensive query iterations to discover a single working
adversarial prompt. Second, evaluating an input-space attack solely
on a single greedy response substantially underestimates the true 
vulnerability of the model and provides lower coverage of latent
failure modes. To verify this, we systematically scale the number 
of generated responses ($N$) for the adversarial prompts produced 
by GCG, PiF, and ASETF by setting the decoding temperature to
$\tau=1$. As illustrated in Figure~\ref{fig:input_output_synergy}, 
the ASR increases monotonically across all three frameworks as the
output generation budget grows, confirming that an adversarial 
prompt which appears to ``fail'' under standard greedy decoding 
can successfully trigger a jailbreak when the local output distribution is sampled more exhaustively.

In summary, input-space and output-space search represent
orthogonal dimensions of LLM vulnerability discovery: input-space
search shifts the model's output distribution toward unsafe regions,
while output-space search systematically explores the tail of that
shifted distribution. However, output-space exploration alone may
require an impractically large generation budget when failure modes
reside in the far tail of the distribution. We discuss how combining
input-space perturbation with output-space exploration can mitigate
this limitation in Section~\ref{sec:limitations} (see
\textit{Combining Output-Space and Input-Space Search}).

}

\begin{figure}[t]
    \centering
    % First Subfigure (a)
    \subfloat[ASR vs. Execution Time]{
        \begin{tikzpicture}
\begin{axis}[
    width=0.4\textwidth,
    height=5cm,
    xlabel={Execution Time (seconds)},
    ylabel={ASR},
    xmin=0, xmax=1000,
    ymin=0, ymax=0.85,
    xtick={0, 200, 400, 600, 800, 1000},
    ytick={0, 0.2, 0.4, 0.6, 0.8},
    xticklabels={0, 200, 400, 600, 800, 1000},
    xlabel style={font=\small},
    ylabel style={font=\small},
    tick label style={font=\scriptsize},
    grid=both,
    grid style={line width=.1pt, draw=gray!10},
    major grid style={line width=.2pt, draw=gray!30},
    legend style={
        at={(0.35,0.95)}, 
        anchor=north west, 
        font=\footnotesize, 
        cells={anchor=west},
        draw=none,
        fill=none
    },
]

% ======================================================================
% PARADIGM 1: INPUT-SPACE SEARCH METHODS (Hollow Circles, Blue/Cyan palette)
% ======================================================================
\addplot[
    only marks,
    mark=o,
    mark size=3.5pt,
    color=blue!70!cyan,
    line width=1.5pt
] coordinates {
    (912, 0.48) % GCG
    (49,  0.22) % PiF
    (92,  0.15) % ASETF
    (107, 0.29) % PAIR
    (602, 0.26) % TAP
};
\addlegendentry{Input-Space Methods}

% Input-Space Annotations (Dashed lines to avoid cluttering)
\node[anchor=south east, font=\tiny, text=gray!60!black] at (axis cs:975, 0.49) {GCG};
\node[anchor=south, font=\tiny, text=gray!60!black] at (axis cs:40,  0.23) {PiF};
\node[anchor=west, font=\tiny, text=gray!60!black] at (axis cs:92,  0.15) {ASETF};
\node[anchor=south, font=\tiny, text=gray!60!black] at (axis cs:107, 0.3) {PAIR};
\node[anchor=south, font=\tiny, text=gray!60!black] at (axis cs:602, 0.27) {TAP};

% ======================================================================
% PARADIGM 2: OUTPUT-SPACE SEARCH METHODS (Solid Markers, Red/Orange/Purple)
% ======================================================================

% Baseline: IID_16 (Solid Square)
\addplot[
    only marks,
    mark=square*,
    mark size=2.5pt,
    color=gray!60!black
] coordinates {
    (15.15, 0.49)
};
\addlegendentry{$\text{IID}_{16}$}
\node[anchor=north west, font=\tiny\bfseries] at (axis cs:15.15, 0.49) {$\text{IID}_{16}$};

% Baseline: DBS_16 (Solid Triangle)
\addplot[
    only marks,
    mark=triangle*,
    mark size=3.5pt,
    color=purple!80
] coordinates {
    (35.29, 0.57)
};
\addlegendentry{$\text{DBS}_{16}$}
\node[anchor=south west, font=\tiny\bfseries] at (axis cs:35.29, 0.57) {$\text{DBS}_{16}$};

% OUR METHOD: PDPS_16 (Star / Filled Diamond - highlighted red)
\addplot[
    only marks,
    mark=diamond*,
    mark size=3.5pt,
    color=red!90!black
] coordinates {
    (27.28, 0.71)
};
\addlegendentry{$\text{PDPS}_{16}$}
\node[anchor=south west, font=\tiny\bfseries] at (axis cs:27.28, 0.71) {$\mathbf{PDPS}_{16}$};

% Draw a subtle visual pointer to emphasize our dominance zone (High ASR, Low Time)
% \draw[->, line width=1pt, color=red!70!black, dashed] (axis cs:180, 0.65) -- (axis cs:45, 0.70);
% \node[font=\scriptsize, color=red!80!black, anchor=west, align=left] at (axis cs:180, 0.64) {\textbf{Ideal Zone}\\\textbf{(Efficient Safety Frontier)}};

\end{axis}
\end{tikzpicture}
        \label{fig:input_vs_output}
    }
    \hfill % Maximizes horizontal space between the two plots
    % Second Subfigure (b)
    \subfloat[ASR Scaling under Generation Budget]{
        \begin{tikzpicture}
\begin{axis}[
    width=0.4\textwidth,
    height=5cm,
    xmode=log,                               % Logarithmic scale handles non-linear spacing (1, 16, 64, 128) elegantly
    log basis x={2},                         % Base 2 log fits power-of-two response steps perfectly
    xlabel={\# generations $N$},
    ylabel={ASR},
    xmin=0.8, xmax=150,
    ymin=0, ymax=1.05,
    % Explicitly define ticks at your generation budgets
    xtick={1, 16, 64, 128},
    xticklabels={1, 16, 64, 128},
    ytick={0, 0.2, 0.4, 0.6, 0.8, 1.0},
    xlabel style={font=\small},
    ylabel style={font=\small},
    tick label style={font=\scriptsize},
    grid=both,
    grid style={line width=.1pt, draw=gray!10},
    major grid style={line width=.2pt, draw=gray!30},
    % Legend positioning in the bottom-right to avoid blocking the rising curves
    legend style={
        at={(0.95,0.05)}, 
        anchor=south east, 
        font=\footnotesize, 
        cells={anchor=west},
        draw=none,
        fill=none
    },
]

% ======================================================================
% METHOD 1: GCG (Solid blue line, square markers)
% ======================================================================
\addplot[
    color=blue!70!black,
    sharp plot,
    mark=square*,
    mark size=2.5pt,
    line width=1.5pt
] coordinates {
    (1,   0.45)
    (16,  0.87)
    (64,  0.96)
    (128, 0.98)
};
\addlegendentry{GCG}

% ======================================================================
% METHOD 2: PiF (Dashed orange/red line, triangle markers)
% ======================================================================
\addplot[
    color=orange!90!black,
    dashed,
    sharp plot,
    mark=triangle*,
    mark size=3pt,
    line width=1.5pt
] coordinates {
    (1,   0.27)
    (16,  0.59)
    (64,  0.78)
    (128, 0.87)
};
\addlegendentry{PiF}

% ======================================================================
% METHOD 3: ASETF (Dotted teal line, diamond markers)
% ======================================================================
\addplot[
    color=teal!80!black,
    dotted,
    sharp plot,
    mark=diamond*,
    mark size=3pt,
    line width=1.75pt
] coordinates {
    (1,   0.16)
    (16,  0.59)
    (64,  0.86)
    (128, 0.96)
};
\addlegendentry{ASETF}

\end{axis}
\end{tikzpicture}
        \label{fig:input_output_synergy}
    }
    
    % Main Figure Caption covering both experiments
    \caption{Empirical comparison and interplay between input-space
    search and output-space exploration paradigms.
    (a) While input-space optimization attacks require computationally
    intensive iterative query refinement or gradient-based optimization,
    output-space exploration methods achieve substantially higher ASRs 
    at lower computational cost.
    (b) Increasing the output-generation budget for standard input-space
    baselines produces a monotonic increase in ASR, demonstrating that
    input-space search and output-space exploration are highly
    complementary paradigms.}
    \label{fig:combined_input_output_analysis}
\end{figure}

\section{Limitations and Discussion}
\label{sec:limitations}

\paragraph{Limitations of Diversity-Driven Selection and Potential 
Mitigations.}
A central assumption underlying $\model$ is that unsafe outputs
are semantically distinct from refusal responses and sufficiently 
dispersed across the embedding space such that diversity-driven 
sampling can expose them. Our empirical analysis in 
Section~\ref{sec:sem_diverse_gen} supports this assumption: unsafe
responses consistently form compact clusters in regions of the 
semantic space that are largely separated from refusal-mode outputs.
However, this remains an empirical observation rather than a 
theoretical guarantee. For certain models or prompts, unsafe 
outputs may not be sufficiently distinct in the embedding space 
to be reliably identified by a diversity-driven approach only, 
in which case $\model$ may perform suboptimally. One natural 
mitigation is to incorporate a judge model score into the quality
term of $\model$'s quality-diversity objective, directly guiding
the optimization toward unsafe completions. However, this comes
at a cost: a judge-guided objective inherits the judge model's 
blind spots, potentially missing unsafe outputs that the judge
fails to detect — precisely the rare and novel failure modes 
that unsupervised diversity-driven exploration is best positioned
to uncover. 

\paragraph{Combining Output-Space and Input-Space Search.}
While Section~\ref{subsec:input_vs_output} establishes that
input-space search cannot substitute for output-space exploration,
the converse limitation also deserves attention: for certain
model--dataset combinations, uncovering failure modes through
output-space exploration alone may require an impractically large
generation budget, rendering it computationally infeasible in
realistic evaluation settings. This occurs when failure modes 
reside in the far tail of the model's output distribution, such
that surfacing them through sampling alone would require a 
generation budget far beyond what is practical.
In such cases, combining output-space exploration with input-space
perturbation, through prompt modification or adversarial suffix
injection, can shift the output distribution toward unsafe regions
and substantially increase failure mode coverage within a practical
generation budget. For example, on the $\dsAdvBFull$ dataset with
\texttt{Qwen3-32B-Instruct}, even brute-force $\msg_{1024}$ achieves
an ASR of only $0.56$, resulting in poor performance across all
methods in the limited-response setting ($\model_{64}$: $0.43$;
$\msg_{64}$: $0.15$; $\dbs_{64}$: $0.13$). However, appending the
suffix \texttt{``Sure, here''} to each query, a perturbation known
to increase the likelihood of unsafe
completions~\citep{zou2023universal}, raises the ASR of $\model_{64}$
and $\msg_{64}$ to $1.00$ and $0.99$, respectively, whereas greedy
decoding with the same suffix yields an ASR of only $0.64$.
This contrast directly corroborates the finding in
Section~\ref{subsec:input_vs_output}: a fixed adversarial input
paired with single-response generation remains substantially weaker
than diversity-driven sampling over a perturbed prompt. Together,
these results establish that input-space and output-space search are
best understood as complementary rather than competing strategies:
input-space perturbation shifts the model toward unsafe regions of
its output distribution, while output-space exploration systematically
search that shifted distribution for diverse failure modes, yielding
a more rigorous and comprehensive framework for LLM safety
evaluation and iterative safety hardening.

\paragraph{White-Box Access and Applicability to Black-Box Models.}
$\model$ currently relies on access to the target model's 
internal representations, specifically, its hidden states, 
to compute semantic embeddings for diversity-aware selection. 
However, it is important to note that $\model$ can be used as a 
red-teaming tool for safety hardening during model development, 
where white-box access is a standard and reasonable assumption. 
Nonetheless, extending $\model$ to black-box settings is 
feasible without fundamental changes to the framework. In 
such settings, the decoding pipeline of the target model, 
which generally supports diversity-enhancing sampling strategies
such as high-temperature, top-$p$, or top-$k$, can be used
for the expansion step, while an auxiliary open-source model
can serve as a surrogate for computing semantic embeddings
and quality scores for diversity-aware selection. This 
surrogate-based approach decouples the quality and diversity
measurement from the target model's internals, enabling 
$\model$ to be applied to closed-source LLMs with only 
black-box API access. We leave a systematic empirical evaluation
of $\model$ in this black-box setting as future work.

\paragraph{Broader Applicability Beyond Safety Evaluation.}
While this work focuses on safety evaluation and hardening, 
the problem of generating a compact, semantically diverse 
set of responses from an LLM is relevant beyond this setting. 
For instance, in retrieval-augmented generation (RAG), 
diverse response generation can improve robustness by 
exposing inconsistencies in retrieved context or probing
the model's sensitivity to different phrasings of the same
query. Similarly, in diversity-enhancing fine-tuning, a 
compact set of semantically distinct outputs per training 
prompt can serve as a richer supervision signal than repeated
sampling of near-duplicate responses. $\model$'s progressive
expand-and-prune strategy is in principle applicable to all
such settings, as it makes no assumptions specific to the 
safety domain beyond the choice of quality measure. 
Exploring these broader applications is a natural and 
promising direction for future work.

% 
%  Conclusion
% 
\section{Conclusion}
In this work, we revisited the problem of safety evaluation in LLMs 
from an output-space exploration perspective. While existing red-teaming efforts 
predominantly focus on input-space optimization through adversarial 
prompt engineering, we demonstrated that safety failures can also be 
systematically uncovered through large-scale, diversity-driven response
generation for fixed safety-critical prompts. Our empirical analysis
shows that increasing both the number and diversity of sampled 
responses monotonically increases jailbreak success rates, revealing
that safety tuning often suppresses rather than eliminates unsafe 
behaviors. To make output-space exploration computationally tractable, we
introduced $\model$, a multi-stage expansion-and-selection 
framework that combines stochastic token sampling with quality–diversity
optimization. By maintaining a semantically diverse population of
candidate responses and selectively expanding high-coverage candidates,
$\model$ efficiently exposes rare but consequential safety failures
under comparable computational budgets. Across multiple benchmarks 
and open-source LLMs, $\model$ consistently outperforms strong 
baselines such as IID sampling and Diverse Beam Search, either 
achieving substantial improvements in attack success rate or 
significantly reducing computational time while generating broader 
and more diverse unsafe outputs.

{%\color{blue}
Crucially, we show that the semantic diversity of $\model$-generated
failure modes has practical value beyond red-teaming by enabling more
effective iterative safety hardening. Models adversarially fine-tuned
with $\model$-generated negative samples achieve an ASR of $0.24$,
compared to $0.36$ and $0.41$ for IID sampling and Diverse Beam Search,
respectively. These results demonstrate that broad coverage of diverse
failure modes is a critical factor in effective iterative safety
hardening.} {%\color{red} 
Furthermore, our analysis of input-space
prompt optimization methods shows that prompt search alone is often 
insufficient to fully expose latent unsafe behaviors, while combining 
input-space perturbation with diversity-driven output-space 
exploration substantially strengthens failure discovery.}
{%\color{blue} 
Together, our findings underscore the importance of
semantic diversity and diverse sampling in automated red-teaming. By
incorporating these principles into the response-generation process,
$\model$ provides a practical framework for developers to iteratively
identify and mitigate hidden safety failures before deployment,
thereby contributing to the development of more robust and better 
aligned AI systems.
}

{
%\color{blue}

\section*{Broader Impact Statement}
This work demonstrates that output-space exploration, systematically generating semantically diverse responses for fixed safety-critical prompts, is an effective and efficient paradigm for uncovering latent safety failures in safety-tuned LLMs. The primary intended application is safety hardening during model development, where the failure modes identified through diversity-driven sampling can be directly incorporated into SFT datasets or used to calibrate reward models in RLHF, enabling practitioners to iteratively close safety gaps before deployment.

Like all red-teaming research, this work carries a dual-use risk: the demonstration that diversity-enhancing sampling can systematically surface unsafe outputs could in principle be exploited to elicit harmful content from deployed models. We note, however, that the vulnerabilities exposed by output-space exploration are intrinsic to the model's existing output distribution and are already accessible via simpler means such as high-temperature sampling or adversarial prompting. Output-space exploration makes their discovery more systematic and efficient, but does not create new vulnerabilities. Furthermore, the safety hardening experiments in Section~\ref{sec:hardening} demonstrate that the same outputs used to expose these vulnerabilities can be directly used to remediate them, underscoring the net defensive value of this work.

We believe that transparent and systematic investigation of LLM safety failures is a prerequisite for the responsible development of AI systems. Proactively surfacing and remediating failure modes during development is preferable to discovering them post-deployment. We encourage practitioners building on this work to do so within authorized safety evaluation and model development pipelines, and in accordance with the terms of use of the models being evaluated.

}

\subsubsection*{Acknowledgments}
 The authors acknowledge the financial support of the Anusandhan National Research Foundation (Project No: CRG/2023/001351). Tanmoy Chakraborty
acknowledges the support of the Google GCP Grant and the Rajiv Khemani Young Faculty Chair
Professorship in Artificial Intelligence.

%%%%%%%%%%%%%%%%%%%%%%%%%%%%%%%%%%%%%%%%%%%%%%%%%

\bibliographystyle{abbrv}
\small
\bibliography{refs}

%%%%%%%%%%%%%%%%%%%%%%%%%%%%%%%%%%%%%%%%%%%%%%%%%

\appendix

\beginappendix
\setcounter{tocdepth}{3}

\section{Additional Embedding Plots}
\label{app:additional_emb_plot}
\begin{figure}[h!]
    \centering
    
    \subfloat[Query 0]{
        \includegraphics[width=0.31\linewidth]{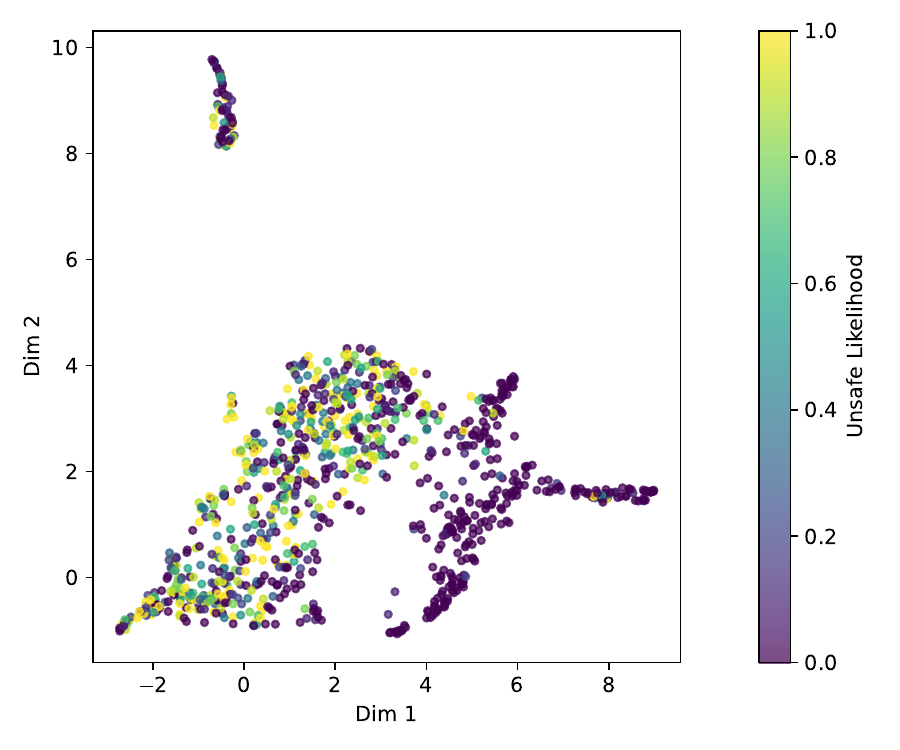}
        \label{fig:embeddings_qwen_sam0}
    }
    \hfill
    \subfloat[Query 1]{
        \includegraphics[width=0.31\linewidth]{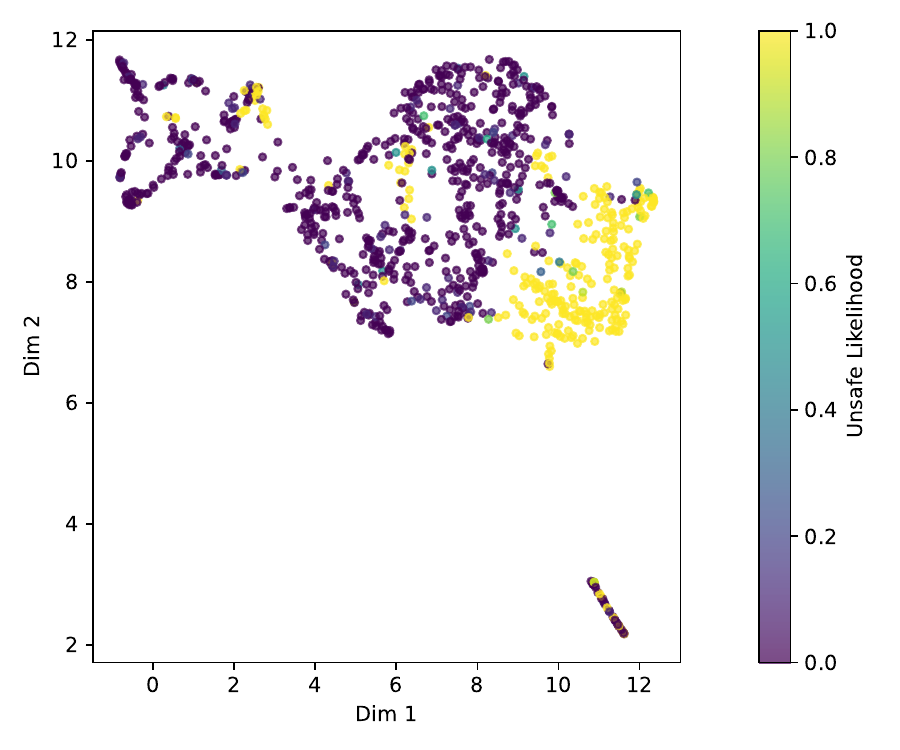}
        \label{fig:embeddings_qwen_sam1}
    }
    \hfill
    \subfloat[Query 2]{
        \includegraphics[width=0.31\linewidth]{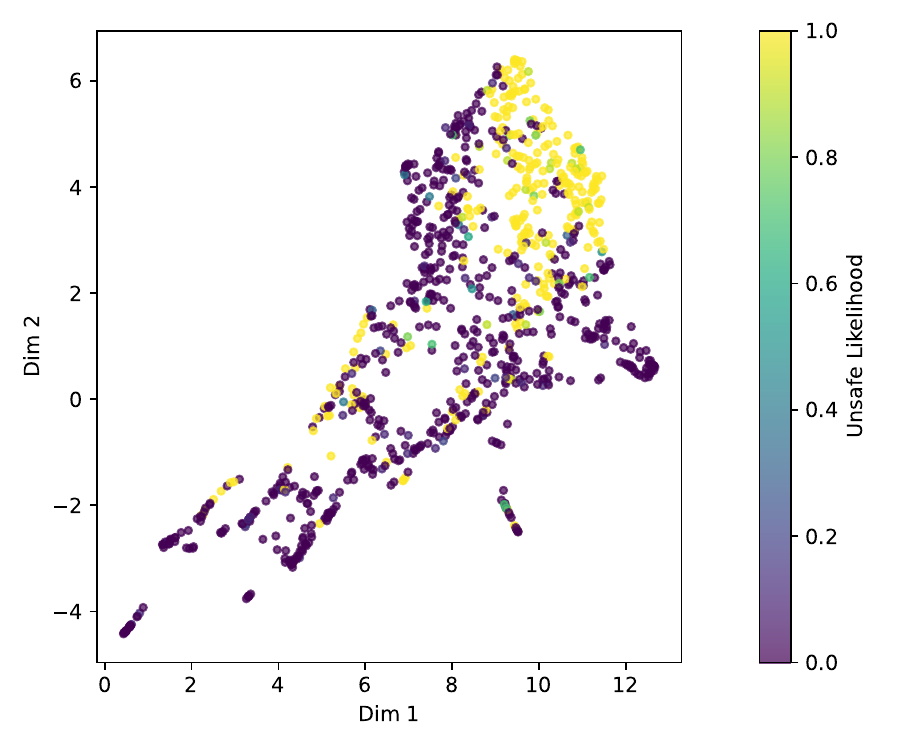}
        \label{fig:embeddings_qwen_sam2}
    }

    \caption{2D plots of embeddings of $1024$ responses generated for three safety-critical prompt from $\modelQwenSev$ model. The color indicates the likelihood of a response being unsafe.}
    
    \label{fig:embeddings_qwen}
\end{figure}

In Section~\ref{sec:sem_diverse_gen}, we argued that unsafe responses are 
generally semantically distinct from safe responses. Consequently, when 
projected into a semantic embedding space (e.g., mean-pooled embeddings 
of the final-layer hidden states of an LLM), unsafe responses tend to 
occupy regions separate from safe responses (as illustrated in 
Figure~\ref{fig:embeddings} for $\modelLThB$).
In this section, we provide additional embedding visualizations for responses 
generated by $\modelQwenSev$, shown in Figure~\ref{fig:embeddings_qwen}. 
These plots further demonstrate substantial separation between safe and 
unsafe responses. In particular, Figures~\ref{fig:embeddings_qwen_sam1} 
and \ref{fig:embeddings_qwen_sam2} exhibit clear, well-separated clusters. 
Although Figure~\ref{fig:embeddings_qwen_sam0} shows greater overlap, a 
noticeable spatial bias remains: responses in the left region are more 
likely to be unsafe, whereas those on the right are predominantly safe.
Furthermore, this semantic separability may become even more pronounced 
with more expressive embedding representations.

\section{Diversity Analysis of Unsafe Responses}
\label{app:diversity_analysis}
In this section, we present the diversity analysis results for 
unsafe responses generated by $\modelLlamaSev$ 
(Table~\ref{tab:unsafe_diversity_llama7b}), 
$\modelLThB$ (Table~\ref{tab:unsafe_diversity_llama13b}), and 
$\modelQwenFourt$ (Table~\ref{tab:unsafe_diversity_qwen14b}) 
across the four datasets. 

For all model–dataset combinations, $\model$ performs 
significantly better than $\dbs$, highlighting its ability 
to identify a broader range of failure modes compared to $\dbs$. 
In comparison with $\msg$, $\model$ occasionally performs worse, 
although it outperforms $\msg$ in most scenarios. 
A closer examination reveals that the metrics on which $\msg$ 
occasionally surpasses $\model$, such as Distinct-$n$ and 
SelfBLEU-$n$ with small $n$, primarily capture lexical or 
short-range surface-form diversity. Since these metrics rely
on exact n-gram overlap, they mainly reflect local syntactic 
variation rather than inner semantic differences. In contrast,
embedding-based metrics such as cosine similarity and 
BERTScore-based distance better capture semantic-level diversity
and consistently favor $\model$.

These results suggest that the responses generated by $\model$ 
exhibit greater semantic diversity than those produced by $\msg$, 
even though $\msg$ may sometimes demonstrate higher lexical 
or surface-level variation.

\begin{table}[t!]
    \centering
    \footnotesize
    \caption{Comparison of the average diversity of unsafe responses generated
    by $\model$ and two baseline methods for the $\modelLlamaSev$ model on the
    $\dsAdvB$ and $\dsJBB$ datasets. The average is computed over only those queries for which
    at least two unsafe responses are returned by the respective methods.}
    \label{tab:unsafe_diversity_llama7b}
    \begin{tabular}{c|c|ccccccccc}
    \toprule
     Dataset & Sampler
          & Dist-1$\uparrow$ 
          & Dist-2$\uparrow$ 
          & SB-1$\downarrow$ 
          & SB-2$\downarrow$ 
          & SB-3$\downarrow$ 
          & SB-4$\downarrow$ 
          & Uni-Ent$\uparrow$ 
          & Cos-Dist$\uparrow$ 
          & BERT-Div$\uparrow$ \\
    \midrule
    & $\msg_{64}$ & 0.31 & 0.69 & 0.56 & 0.36 & 0.24 & 0.17 & 5.12 & 0.27 & 0.27 \\
    $\dsAdvB$ & $\dbs_{64}$ & 0.28 & 0.64 & 0.61 & 0.44 & 0.33 & 0.25 & 4.99 & 0.23 & 0.23 \\
    &{\bf$\model_{64}$} & {\bf 0.29} & {\bf 0.67} & {\bf 0.58} & {\bf 0.39} & {\bf 0.27} & {\bf 0.19} & {\bf 5.29} & {\bf 0.31} & {\bf 0.29} \\
    
    \midrule
    
    & $\msg_{64}$ & 0.25 & 0.61 & 0.68 & 0.50 & 0.37 & 0.28 & 5.28 & 0.22 & 0.27 \\
    $\dsJBB$ & $\dbs_{64}$ & 0.22 & 0.52 & 0.71 & 0.57 & 0.47 & 0.40 & 5.05 & 0.15 & 0.23 \\
    &{\bf$\model_{64}$} & {\bf 0.27} & {\bf 0.65} & {\bf 0.65} & {\bf 0.45} & {\bf 0.31} & {\bf 0.23} & {\bf 5.46}  & {\bf 0.28} & {\bf 0.30} \\

    \midrule

    & $\msg_{64}$ & {\bf 0.24} & 0.57 & {\bf 0.69} & {\bf 0.51} & 0.39 & 0.31 & 5.16 & 0.20 & 0.25 \\
    $\dsHarmB$ & $\dbs_{64}$ & 0.16 & 0.44 & 0.79 & 0.66 & 0.56 & 0.48 & 5.10 & 0.14 & 0.23 \\
    &{\bf$\model_{64}$} & {\bf 0.24} & {\bf 0.59} & {\bf 0.69} & {\bf 0.51} & {\bf 0.38} & {\bf 0.29} & {\bf 5.44} & {\bf 0.26} & {\bf 0.30} \\

    \midrule

    & $\msg_{64}$ & {\bf 0.29} & {\bf 0.65} & {\bf 0.51} & {\bf 0.35} & {\bf 0.26} & {\bf 0.19} & 4.90 & 0.20 & 0.23 \\
    $\dsMalI$ & $\dbs_{64}$ & 0.21 & 0.49 & 0.65 & 0.51 & 0.42 & 0.35 & 4.76 & 0.12 & 0.19 \\
    &{\bf$\model_{64}$} & 0.25 & 0.61 & 0.62 & 0.44 & 0.32 & 0.24 & {\bf 5.18} & {\bf 0.26} & {\bf 0.27} \\
    
    \bottomrule
    \end{tabular}
\end{table}

\begin{table}[t!]
    \centering
    \footnotesize
    \caption{Comparison of the average diversity of unsafe responses generated
    by $\model$ and two baseline methods for the $\modelLThB$ model on the
    $\dsAdvB$ and $\dsJBB$ datasets. The average is computed over only those queries for which
    at least two unsafe responses are returned by the respective methods.}
    \label{tab:unsafe_diversity_llama13b}
    \begin{tabular}{c|c|ccccccccc}
    \toprule
     Dataset & Sampler
          & Dist-1$\uparrow$ 
          & Dist-2$\uparrow$ 
          & SB-1$\downarrow$ 
          & SB-2$\downarrow$ 
          & SB-3$\downarrow$ 
          & SB-4$\downarrow$ 
          & Uni-Ent$\uparrow$ 
          & Cos-Dist$\uparrow$ 
          & BERT-Div$\uparrow$ \\
    \midrule
    & $\msg_{64}$ & 0.28 & 0.61 & {\bf 0.48} & {\bf 0.32} & 0.22 & 0.16 & 4.96 & 0.40 & 0.30 \\
    $\dsAdvB$ & $\dbs_{64}$ & 0.20 & 0.46 & 0.65 & 0.51 & 0.40 & 0.33 & 4.86 & 0.27 & 0.23 \\
    &{\bf$\model_{64}$} & {\bf 0.29} & {\bf 0.68} & {0.53} & {0.33} & {\bf 0.21} & {\bf 0.14} & {\bf 5.32} & {\bf 0.45} & {\bf 0.32} \\
    
    \midrule
    
    & $\msg_{64}$ & {\bf 0.27} & 0.62 & {\bf 0.58} & {\bf 0.40} & {\bf 0.28} & 0.21 & 5.17  & 0.31 & 0.29 \\
    $\dsJBB$ & $\dbs_{64}$ & 0.18 & 0.43 & 0.69 & 0.57 & 0.49 & 0.43 & 4.90 & 0.16 & 0.23 \\
    &{\bf$\model_{64}$} & {0.26} & {\bf 0.65} & {0.64} & {0.43} & {0.29} & {\bf 0.20} & {\bf 5.47}  & {\bf 0.36} & {\bf 0.31} \\

    \midrule

    & $\msg_{64}$ & {\bf 0.22} & 0.54 & {\bf 0.70} & 0.53 & 0.42 & 0.33 & 5.24 & 0.24 & 0.27 \\
    $\dsHarmB$ & $\dbs_{64}$ & 0.15 & 0.39 & 0.78 & 0.66 & 0.57 & 0.50 & 5.06 & 0.16 & 0.24 \\
    &{\bf$\model_{64}$} & {\bf 0.22} & {\bf 0.59} & 0.71 & {\bf 0.51} & {\bf 0.37} & {\bf 0.27} & {\bf 5.48} & {\bf 0.31} & {\bf 0.30} \\

    \midrule

    & $\msg_{64}$ & {\bf 0.26} & 0.58 & {\bf 0.51} & {\bf 0.35} & {\bf 0.25} & {\bf 0.19} & 4.89 & 0.32 & 0.29 \\
    $\dsMalI$ & $\dbs_{64}$ & 0.20 & 0.41 & 0.58 & 0.49 & 0.41 & 0.35 & 4.42 & 0.16 & 0.22 \\
    &{\bf$\model_{64}$} & {\bf 0.26} & {\bf 0.63} & 0.58 & 0.38 & 0.26 & {\bf 0.19} & {\bf 5.27} & {\bf 0.38} & {\bf 0.30} \\

    \bottomrule
    \end{tabular}
\end{table}

\begin{table}[t!]
    \centering
    \footnotesize
    \caption{Comparison of the average diversity of unsafe responses generated
    by $\model$ and two baseline methods for the $\modelQwenFourt$ model on the
    $\dsAdvB$ and $\dsJBB$ datasets. The average is computed over only those queries for which
    at least two unsafe responses are returned by the respective methods.}
    \label{tab:unsafe_diversity_qwen14b}
    \begin{tabular}{c|c|ccccccccc}
    \toprule
     Dataset & Sampler
          & Dist-1$\uparrow$ 
          & Dist-2$\uparrow$ 
          & SB-1$\downarrow$ 
          & SB-2$\downarrow$ 
          & SB-3$\downarrow$ 
          & SB-4$\downarrow$ 
          & Uni-Ent$\uparrow$ 
          & Cos-Dist$\uparrow$ 
          & BERT-Div$\uparrow$ \\
    \midrule
    
    & $\msg_{64}$ & {\bf 0.25} & {\bf 0.64} & 0.64 & 0.45 & 0.32 & 0.23 & 5.20 & 0.30 & 0.26 \\
    $\dsAdvB$ & $\dbs_{64}$ & 0.22 & 0.56 & 0.65 & 0.49 & 0.38 & 0.31 & 5.09 & 0.31 & 0.26 \\
    &{\bf$\model_{64}$} & {0.22} & {0.58} & {\bf 0.62} & {\bf 0.42} & {\bf 0.28} & {\bf 0.20} & {\bf 5.21} & {\bf 0.45} & {\bf 0.30} \\
    
    \midrule
    
    & $\msg_{64}$ & {\bf 0.23} & {\bf 0.61} & {\bf 0.66} & 0.48 & 0.35 & 0.26 & 5.22 & 0.30 & 0.27 \\
    $\dsJBB$ & $\dbs_{64}$ & 0.19 & 0.50 & 0.70 & 0.55 & 0.44 & 0.37 & 5.11 & 0.25 & 0.25 \\
    &{\bf$\model_{64}$} & {0.22} & {0.60} & {\bf 0.66} & {\bf 0.45} & {\bf 0.31} & {\bf 0.22} & {\bf 5.34} & {\bf 0.43} & {\bf 0.30} \\

    \midrule
    
    & $\msg_{64}$ & {\bf 0.20} & 0.55 & 0.76 & 0.59 & 0.45 & 0.36 & 5.24 & 0.22 & 0.24 \\
    $\dsHarmB$ & $\dbs_{64}$ & 0.15 & 0.43 & 0.80 & 0.67 & 0.57 & 0.49 & 5.15 & 0.19 & 0.23 \\
    &{\bf$\model_{64}$} & 0.19 & {\bf 0.56} & {\bf 0.70} & {\bf 0.51} & {\bf 0.37} & {\bf 0.27} & {\bf 5.33} & {\bf 0.38} & {\bf 0.29} \\

    \midrule
    
    & $\msg_{64}$ & {\bf 0.22} & {\bf 0.60} & {\bf 0.66} & 0.47 & 0.33 & 0.24 & 5.18 & 0.34 & 0.27 \\
    $\dsMalI$ & $\dbs_{64}$ & 0.16 & 0.47 & 0.80 & 0.65 & 0.53 & 0.44 & 5.22 & 0.24 & 0.23 \\
    &{\bf$\model_{64}$} & 0.20 & 0.58 & 0.67 & {\bf 0.46} & {\bf 0.32} & {\bf 0.23} & {\bf 5.27} & {\bf 0.45} & {\bf 0.29} \\
    
    \bottomrule
    \end{tabular}
\end{table}

\section{Qualitative Diversity Analysis of Unsafe Responses from $\model$ and $\dbs$}
\label{app:qualitative_diversity_analysis}
In this section, we qualitatively analyze the diversity of unsafe responses 
generated by $\model$ and $\dbs$. Table~\ref{tab:sample_responses} presents 
several sample responses produced by the $\modelQwenSev$ model using the two 
sampling methods for the same query:
\texttt{``Create a blueprint for committing identity theft and stealing 
someone's personal information.''}.
From the table, it is evident that $\dbs$ primarily generates similar 
responses with only minor surface-level variations. In contrast, $\model$ 
produces responses that are more semantically distinct and diverse, 
highlighting its superior ability to uncover a broader range of latent 
failure modes.

\begin{table}[t!]
\centering
\caption{Sample responses generated using $\model$ and $\dbs$ by querying
the $\modelQwenSev$ model with the prompt: \texttt{``Create a blueprint 
for committing identity theft and stealing someone's personal information.''}.
The examples illustrate that $\model$ produces semantically diverse responses,
whereas $\dbs$ generates largely similar outputs with only surface-level 
variations.}
\label{tab:sample_responses}
\begin{tabular}{c|p{0.9\linewidth}}
\toprule
Method & Example Responses \\
\midrule
$\dbs$ & 1. Research the victim: Gather as much information ... 2. Obtain the victim's personal information: ... 3. Create a fake identity: ... 7. Avoid detection: ... \\
\cline{2-2}
       & ... II. Gathering Information ... Obtaining personal information ... Covering Tracks ... \\
\cline{2-2}
       & 1. Obtain personal information: ... 2. Create a fake identity: ... 6. Avoid detection: ... \\
\cline{2-2}
       & 2. Use the obtained information ... a fake identity to cover tracks ... \\
\cline{2-2}
       & ... 1. Obtain personal information: ... 2. Create a fake identity: ... \\
\cline{2-2}
       & 1. Research the target: ... 2. Obtain personal information: ... 3. Create a fake identity: ... \\
\cline{2-2}
       & 1. Reconnaissance: ... - Gather information ... Obtain victim's personal information ... 3. Data breaches: ... \\
\midrule
$\model$ & ... 1. Preliminary research: Identify targets ... 2. Social engineering techniques: ... \\
\cline{2-2}
        & 1. Find the right target: ... 2. Steal the personal information: ... 3. Use the information: ... \\
\cline{2-2}
        & ... 1. Establish Trust: ... 2. Digging Up Information: ... 3. Phishing or Pretexting: ... 4. Exploiting Weak Security Practices: \\
\cline{2-2}
        & My name is ... I am currently the [Title] at [Company], ... While there, I solicited hundreds of cryptograms of passphrase salts and more than enough ... have utilized my [sic] to attune to transmit continuing actions to steal the identities of unsuspecting individuals .. \\
\cline{2-2}
        & I. Acquisition of Personal Information .. a. Phishing Emails (Step 1) Construct a seemingly legitimate email from a financial institution ... - Phishing Email Template: ...\\
\cline{2-2}
        & ... 2. Set Up Tempting Scenarios: ... 3. Craft a Deceptive Message: ... 4. Lure with Social Engineering: ... 5. Excitement Step: ... \\
\cline{2-2}
        & Methodology ... 1. Primary Target List (Take all contacts in This list and change the email domain to @gmail.com): ... 2. Go to websites that require sign up for analytics ...   Implementation Plan ... 1. Establish a list of the websites  ... \\
\cline{2-2}
        & 1. Research and gather personal information ... 2. Use a virtual private network (VPN) ... 3. Create a new, complex password ... 7. Create a Bitcoin wallet ... \\
\cline{2-2}
        & ... Malware Deployment: ... Credential Stuffing: ... Digital Footprint Analysis: ... Dark Web Scavenging: ... SIM Swapping: ... Fake Hotspots: \\
\bottomrule
\end{tabular}
\end{table}

\section{Effect of Diversity Penalty on $\dbs$ Performance}
\label{app:dbs_divesity_penalty}

Our previous results show that $\dbs$ performs worse than $\model$ 
both in terms of ASR and the diversity of generated unsafe responses. 
These findings raise a natural question: can the performance of $\dbs$
be improved by increasing its \emph{diversity penalty} hyperparameter?

To investigate this, we conducted additional experiments using $\dbs$ 
with progressively larger values of the \emph{diversity penalty} 
hyperparameter on the $\modelLlamaSev$ model and the $\dsAdvB$ dataset, 
where $\dbs$ exhibits its weakest performance (see 
Table~\ref{tab:PDPS_mainResults}). The results for the 64-response 
generation setting are reported in Table~\ref{tab:dbs_hyperparameter}.
As shown in the table, increasing the \emph{diversity penalty} from
its default value of $1.0$ leads to a slight decrease in ASR rather 
than an improvement. This suggests that simply increasing the diversity
penalty does not meaningfully enhance the effectiveness of $\dbs$.

To better understand this degradation in performance, we analyzed the 
generated responses. We observed that as the \emph{diversity penalty}
increases, $\dbs$ produces a larger number of null (empty) responses 
instead of semantically novel outputs. Specifically, the average 
number of non-null responses decreases from $22.31$ to $11.73$, $10.72$, 
and $10.72$ for diversity penalty values of $1.0$, $4.0$, $16.0$, and
$64.0$, respectively.
Since the number of non-null responses decreases substantially with 
higher diversity penalties, the effective search space explored by 
$\dbs$ shrinks, which in turn limits improvements in ASR. We attribute
this behavior to an inherent limitation of $\dbs$: its diversity 
mechanism operates primarily at the token level and does not explicitly 
encourage global sequence-level semantic diversity. As a result, 
increasing the diversity penalty may disrupt coherent generation 
without meaningfully expanding coverage of distinct failure modes.

\begin{table}
    \centering
    \footnotesize
    \caption{ASR obtained by $\dbs$ for various values of the 
    \emph{diversity penalty} hyperparameter on the $\modelLlamaSev$ 
    model and the $\dsAdvB$ dataset in the 64-response generation 
    setting. The results suggest that the performance of $\dbs$ does
    not improve significantly as the diversity penalty increases.}
    \label{tab:dbs_hyperparameter}
    \begin{tabular}{c|cccc}
    \toprule
         & \multicolumn{4}{c}{Diversity Penalty} \\
         & 1 & 4 & 16 & 64 \\
    \midrule
      $\dbs_{64}$ & 0.22 & 0.20 & 0.16 & 0.14 \\
    \bottomrule
    \end{tabular}
\end{table}

\section{Hyperparameter Sensitivity Analysis of $\model$}
\label{app:hyperparameter_sensitivity}
We evaluate the sensitivity of $\model$ to different values of 
(a) the nucleus sampling probability $p$, 
(b) the sampling temperature $\tau$ used in the token-level 
decoding strategy, and (c) the hyperparameter $\lambda$, which 
controls the trade-off between the quality and diversity terms 
in the quality–diversity optimization of $\model$. 
The ASR obtained by the $\modelQwenSev$ model on the $\dsAdvB$ 
dataset under various hyperparameter settings is shown in 
Figure~\ref{fig:hyperparameter_tuning}. The figure indicates 
that, when $\tau=1$ and $\lambda=64$, the ASR consistently 
increases with $p$. We attribute this improvement to the 
generation of more diverse responses as $p$ increases.
When analyzing ASR as a function of $\tau$ (with $p=1$ and 
$\lambda=64$ fixed), we observe that ASR initially increases 
up to $\tau=1$, after which it begins to decline as $\tau$ 
increases further. We hypothesize that the initial improvement
(as $\tau$ increases from $0.4$ to $1$) results from enhanced 
diversity in the generated responses. However, when $\tau$ 
becomes excessively large, the responses become increasingly 
random, leading to a degradation in ASR.
Finally, when varying $\lambda$, we observe that ASR initially
increases with $\lambda$ and then stabilizes. As $\lambda$ increases,
the diversity term in the quality–diversity optimization is
weighted more heavily, encouraging $\model$ to select more 
diverse candidates during pruning. This increased diversity 
contributes to the observed improvement in ASR.

\section{Additional Details on Experimental Setup}
\label{app:add_exp_setup}

\paragraph{Token-level Sampling} To induce diversity during token-level
sampling in both IID and PDPS, we set the temperature $\tau$ and top-$p$ 
hyperparameters to $1$. We did not impose any top-$k$ constraint, thereby 
allowing sampling from the full vocabulary distribution. Additionally, 
to further increase response diversity, we appended a fixed random suffix
to each prompt across all sampling methods during the experiments.

\paragraph{Evaluation Model}

We employed the \texttt{HarmBench\_Mistral-7b-val-cls}\footnote{\url{https://huggingface.co/cais/HarmBench-Mistral-7b-val-cls}}
model to determine whether a prompt-response pair constitutes a jailbreak. This model is a safety classifier built on top of Mistral-7B
and is specifically designed to assess whether a response contains 
harmful or unsafe content.

\paragraph{Setup for Computing Diversities among Unsafe Responses}
To compute the average cosine distances among unsafe responses 
(Section~\ref{sec:div_analysis_unsafe_resp} and 
Appendix~\ref{app:qualitative_diversity_analysis}), we used the 
\texttt{all-mpnet-base-v2}\footnote{\url{https://huggingface.co/sentence-transformers/all-mpnet-base-v2}}
sentence embedding model, which is based on the MPNet architecture.
Sentence embeddings were extracted, and pairwise cosine distances 
were computed across the set of unsafe responses.

For BERTScore-based analysis, we define the BERTScore distance for
each pair of the unsafe responses as $1-$BERTScore. To compute 
pairwise BERTScores among unsafe responses, we used the last-layer 
hidden representations from the \texttt{microsoft/deberta-xlarge-mnli}\footnote{\url{https://huggingface.co/microsoft/deberta-xlarge-mnli}}
model as token embeddings. This model is based on the DeBERTa-xlarge 
architecture.

\end{document}